\documentclass{article}
\newif\ifconferenceformat
\IfFileExists{iclr2027_conference.sty}{
  \conferenceformattrue
  \usepackage{iclr2027_conference,times}
}{
  \conferenceformatfalse
  \usepackage[letterpaper,textwidth=5.5in,textheight=9in,centering]{geometry}
  \usepackage{times,natbib,fancyhdr}
  \setlength{\parindent}{0pt}
  \setlength{\parskip}{5pt}
  \pagestyle{fancy}\fancyhf{}
  \fancyhead[C]{\scriptsize EDITORIAL WORKING DRAFT -- NON-OFFICIAL LAYOUT}
  \fancyfoot[C]{\thepage}
  \setlength{\headheight}{14pt}
  \fancypagestyle{plain}{\fancyhf{}\fancyhead[C]{\scriptsize EDITORIAL WORKING DRAFT -- NON-OFFICIAL LAYOUT}\fancyfoot[C]{\thepage}}
  \PackageWarningNoLine{VisProve}{Official ICLR 2027 style absent. Preview only}
}
\usepackage[T1]{fontenc}
\usepackage[utf8]{inputenc}
\ifconferenceformat\else\date{}\fi

\usepackage{amsmath,amsfonts}

\usepackage{graphicx}
\usepackage{float}
\usepackage{hyperref}
\usepackage{xurl}
\usepackage{booktabs}
\usepackage{array}
\usepackage{amssymb}
\usepackage{wrapfig}
\usepackage{multirow}
\usepackage[table]{xcolor}
\usepackage{makecell}
\usepackage{colortbl}
\usepackage[most]{tcolorbox}
\tcbuselibrary{breakable,skins}
\usepackage{capt-of}

\hypersetup{hidelinks}
\definecolor{prompttitle}{RGB}{132,180,225}
\definecolor{promptbg}{RGB}{242,247,252}
\definecolor{promptframe}{RGB}{160,195,225}

\newtcolorbox{promptfigure}[1]{
    enhanced,
    breakable,
    colback=promptbg,
    colframe=promptframe,
    colbacktitle=prompttitle,
    coltitle=white,
    title={#1},
    fonttitle=\bfseries,
    boxrule=0.5pt,
    arc=1.2mm,
    left=2mm,
    right=2mm,
    top=1.5mm,
    bottom=1.5mm,
    before skip=0.8em,
    after skip=0.4em
}
\definecolor{bestbg}{RGB}{249,235,221}
\definecolor{secondbg}{RGB}{233,244,228}
\newcommand{\best}[1]{\cellcolor{bestbg}\textbf{#1}}
\newcommand{\second}[1]{\cellcolor{secondbg}#1}

\title{VisProve: Learning Typed Visual Provenance for Scientific Document Question Answering}

\author{%
Jingxuan Wei\textsuperscript{1,2,3},
Zhengyang Wang\textsuperscript{2},
Zheng Guo\textsuperscript{2},
Xiaoming Wu\textsuperscript{2},\\[0.2em]
\textbf{Fuqiang Wang\textsuperscript{2,3},
Xin Wang\textsuperscript{2}\footnotemark[1],
Bihui Yu\textsuperscript{1},
Yining Wang\textsuperscript{4},
Junnan Zhu\textsuperscript{5}\thanks{Corresponding authors.}}\\[0.6em]
\textsuperscript{1}\,University of Chinese Academy of Sciences\\
\textsuperscript{2}\,Qilu University of Technology\\
\textsuperscript{3}\,Shandong Computer Science Center\\
\textsuperscript{4}\,Unisound AI\\
\textsuperscript{5}\,Institute of Automation, Chinese Academy of Sciences\\[0.4em]
\href{mailto:junnan.zhu@nlpr.ia.ac.cn}{\texttt{junnan.zhu@nlpr.ia.ac.cn}}\\
\href{mailto:weijingxuan20@mails.ucas.edu.cn}{\texttt{weijingxuan20@mails.ucas.edu.cn}}\quad
\href{mailto:wangzhengyang291@gmail.com}{\texttt{wangzhengyang291@gmail.com}}
}

\ifconferenceformat
  \iclrfinalcopy
\fi

\begin{document}
\raggedbottom
\maketitle
\ifconferenceformat
  \lhead{Preprint}
\fi

\begin{abstract}
Visual citations show where an answer is supported, but not how the cited content contributes to it. In scientific document question answering, a model may read a value directly, compress several observations, or combine evidence to infer a conclusion. We study typed visual provenance, where a system generates an answer together with source regions and explicit support relations. We introduce \textbf{ReViE}, a dataset of 3,941 questions over complete scientific documents, with page, section, and region annotations labeled as \emph{Quotation}, \emph{Compression}, or \emph{Inference}. We further propose \textbf{VisProve}, which augments joint answer and provenance generation with differentiable Spatial Evidence Reading and Evidence--Answer Relation Feedback. VisProve conditions next-token prediction on localized visual evidence and its estimated support relation through a gated norm-bounded residual. On the document-disjoint ReViE test set, VisProve improves Typed-F1 from 3.52 to 5.48 and Section-F1 from 45.99 to 53.24 over joint supervised fine-tuning with the same backbone, while Page-F1 changes only marginally. These results suggest that coupling visual grounding with support relation modeling mainly improves fine-grained provenance, while the low absolute Typed-F1 reveals a substantial gap between answer generation and precise visual evidence attribution.

\end{abstract}

\section{Introduction}
\label{sec:introduction}

An answer to a scientific question is easier to assess when its evidence is visible. However, locating the correct table, figure, or text region does not fully explain how that evidence supports the answer. A model may read a single entry directly, summarize several observations, or compare multiple regions to derive a conclusion. These cases require different forms of verification. A bounding box therefore identifies where to look, but not how the cited region contributes to the answer. This distinction is especially important in scientific document question answering, where text, figures, and tables may provide complementary evidence across pages~\citep{mmlongbenchdoc,sciegqa}.

Existing visual grounding benchmarks make supporting pages and regions inspectable~\citep{sciegqa,citevqa,m3grounder}. In parallel, textual provenance research distinguishes how a source is used, including direct restatement, compression, and inference~\citep{trove,genprove}. These two lines of work capture complementary aspects of provenance, but they are usually modeled separately. We ask whether explicitly modeling both visual source locations and support relations can improve provenance beyond learning to serialize answers and citations jointly. Such relation labels make the proposed use of evidence explicit and evaluable, while remaining distinct from a claim that the evidence itself is correct.

We study \emph {fine-grained typed visual provenance}. Given a question and the complete ordered sequence of document page images, a model generates an answer and a set of evidence items. Each item specifies a page, a section, a bounding box, and a support relation. We use \emph{Quotation}, \emph{Compression}, and \emph{Inference} to distinguish information that is directly available in a region, information that is faithfully condensed from the evidence, and conclusions that require reasoning over the evidence. Fine granularity refers to the localization of regions rather than to sentence or claim decomposition of the answer. Spatial extent and semantic use are treated separately. For example, a directly stated sentence that spans two pages remains Quotation even though its evidence occupies multiple regions.

To support this task, we construct \textbf{ReViE} from SciEGQA~\citep{sciegqa}. ReViE combines refined question-answering pairs from SciEGQA with newly constructed, visually verified multi-hop questions and a separate set of cross-page-contiguous questions. The dataset contains 3,941 instances linked to 8,523 visual evidence regions, with page, section, region, and support relation annotations under document-disjoint splits. We restore the complete ordered page sequence for every document, so models must locate sparse supporting evidence within the full document rather than reason only over preselected gold evidence.

We further propose \textbf{VisProve}, which couples evidence reading with support relation modeling during generation. Joint supervised fine-tuning (SFT) learns answer content and structured provenance, but it does not explicitly connect a predicted location to a differentiable representation of the visual content at that location. \emph{Spatial Evidence Reading} (SER) addresses this gap by constructing an evidence representation from a page distribution and a latent spatial mask, both conditioned on the question and the current decoder state. \emph{Evidence--Answer Relation Feedback} (ERF) maps this evidence representation to a distribution over support relations. VisProve then integrates the evidence and relation representations through a gated norm-bounded residual before the original language head. The model continues to generate both answers and provenance fields autoregressively, and its internal location and relation estimates guide generation without replacing the emitted citations.

Our main comparison is against joint supervised fine-tuning using the same backbone and training split. On the document-disjoint ReViE test set, VisProve improves Typed-F1 from 3.52 to 5.48 and Section-F1 from 45.99 to 53.24, while Page-F1 changes from 75.00 to 75.73. The larger gains at the section and typed levels suggest that the benefit is concentrated in finer provenance rather than coarse page localization. Feedback ablations also favor the full configuration for typed provenance, although improvements in answer quality are not uniform across metrics. Overall, these results suggest that explicit evidence reading and relation modeling provide complementary signals, while the low absolute Typed-F1 highlights the remaining difficulty of precise typed visual provenance.

\begin{figure}[t]
    \centering
    \includegraphics[width=\linewidth]{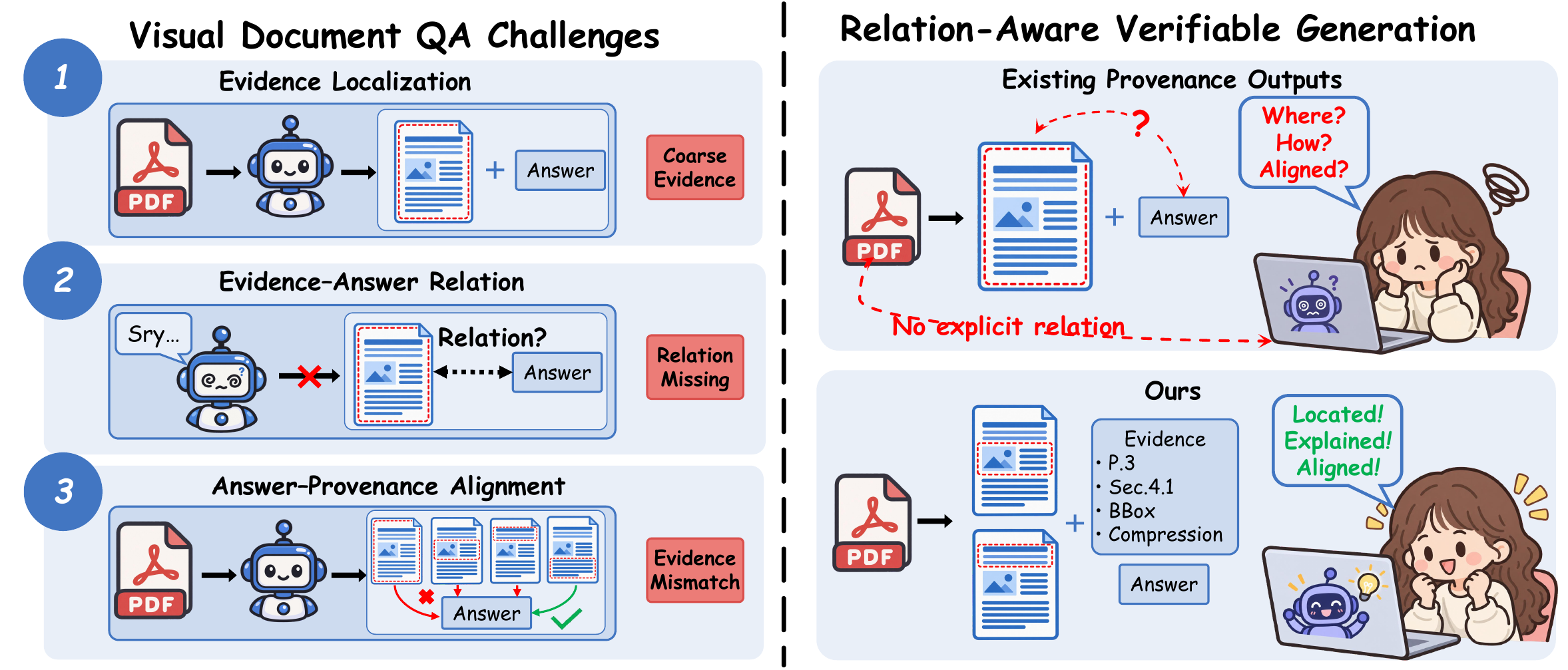}
    \caption{Overview of fine-grained typed visual provenance. The model links a complete answer to supporting visual regions and specifies how each region contributes to the answer.}
    \label{fig:motivation}
\end{figure}

Our contributions are summarized as follows.
\begin{itemize}
    \item We introduce ReViE, which combines hierarchical visual source localization with region-level support relations under complete document input.
    \item We propose VisProve, which connects differentiable spatial evidence reading and support relation feedback to joint answer and provenance generation.
    \item We provide shared-backbone comparisons against joint SFT, quantify gains in fine-grained provenance, and expose the substantial remaining gap in precise typed visual provenance.
\end{itemize}


\section{Related Work}
\label{sec:related_work}

\textbf{Visual document question answering.}
Document QA has progressed from OCR- and layout-aware representations~\citep{docvqa,layoutlmv2,tilt} to direct understanding of document images~\citep{donut,pix2struct,ureader}. Multi-page models address cross-page aggregation and visual compression~\citep{gram,mplugdocowl2}, while long-document benchmarks examine understanding and reasoning over extended visual contexts~\citep{mmlongbenchdoc,longdocurl}. Retrieval, interactive navigation, and multi-step reasoning offer complementary strategies for accessing relevant evidence~\citep{vdocrag,docvstar,mmdocr1,trace}. UniDoc-RL~\citep{unidocrl} combines retrieval, reranking, active visual perception, and reasoning within a reinforcement-learning framework, progressively narrowing evidence from documents to images and local regions. Complementing these evidence-acquisition pipelines, SCVER~\citep{scver} enables decoder-state-dependent access to fine-grained visual features during autoregressive document parsing. Our setting takes the complete, ordered page-image sequence of a scientific document as input and requires an answer together with structured visual provenance, making evidence localization part of the output rather than only an intermediate processing step.

\textbf{Visual evidence grounding and provenance.}
Visual grounding associates answers with spatial support, including page references and bounding boxes~\citep{visa,dogr,citevqa}. SciEGQA provides semantic-region evidence for scientific-document QA~\citep{sciegqa}, while M3Grounder formulates multi-span and multi-granular grounding as pixel-level segmentation~\citep{m3grounder}. Grounding supervision also supports evidence-aware chart and document reasoning~\citep{refchartqa,docr1}. Beyond localization alone, LAT~\citep{lat} and Chain of Evidence~\citep{chainofevidence} integrate attribution into reasoning or generation. DocSeeker~\citep{docseeker} organizes document reasoning into analysis, evidence localization, and reasoning, while CoExVQA~\citep{coexvqa} explicitly localizes answer regions and restricts answer decoding to the grounded content. VinQA~\citep{vinqa} interleaves cited visual elements with long-form answers, and DocTrace~\citep{doctrace} builds hierarchical evidence graphs with node-level provenance. More recently, DocAttriBench~\citep{docattribench} constructs element-level grounding annotations using changes in answer perplexity after masking candidate document elements. VisProve focuses on jointly predicting hierarchical visual locations and the semantic support relation of each evidence region to the complete answer, while using relation feedback to condition generation.

\textbf{Textual provenance and evidence--answer relations.}
Evidence-based text generation studies how generated content can be connected to supporting sources, with diverse choices of attribution granularity and evaluation~\citep{evidencegensurvey}. Sentence- and span-level attribution provide localized source links~\citep{longcite,laquer}, while ALiiCE examines fine-grained citation placement within generated sentences~\citep{aliice}. Beyond identifying sources, TROVE annotates evidence-use relations, including \emph{Quotation}, \emph{Compression}, and \emph{Inference}~\citep{trove}. GenerationPrograms makes source transformations explicit through executable text operations~\citep{generationprograms}, and GenProve jointly generates answers and sentence-level textual provenance with semantic support relations~\citep{genprove}. Extending this perspective to multimodal agents, TRACER links generated sentences to supporting tool observations and their evidence-use relations~\citep{tracer}. ReViE adopts relation-aware provenance for native scientific-document images while retaining complete-answer supervision: a complete answer is linked to a set of visual evidence regions rather than decomposed into individually attributed sentences. Each evidence item specifies a page, section, bounding box, and relation. The resulting task couples locating visual support with identifying whether it supplies directly stated information, a faithful condensation, or an inference supporting the answer.

\section{ReViE: Relation-aware Visual Evidence}
\label{sec:dataset}

\subsection{Task and Annotation Schema}
\label{sec:dataset_overview}

Given a question $q$ and an ordered sequence of document page images $D=(I_1,\ldots,I_M)$, the task is to generate an answer $a$ and an evidence set $E=\{(p_k,s_k,b_k,r_k)\}_{k=1}^{K}.$

Each item records its 1-based page index $p_k$, section or subsection title $s_k$, bounding box $b_k$, and relation $r_k\in\{\textit{Quotation}, \textit{Compression}, \textit{Inference}\}$. The serialized fields are \texttt{page}, \texttt{section}, \texttt{rel\_bbox}, and \texttt{relation}; box coordinates use a 0--1000 page-relative scale, and an unresolved section uses JSON \texttt{null}.

The evidence set supports the complete answer. Each region is labeled according to the answer content it contributes, and multiple regions can jointly support a conclusion. \emph{Quotation} covers direct reading and minor rephrasing; \emph{Compression} covers condensation of explicitly available information; and \emph{Inference} covers conclusions requiring comparison, calculation, or reasoning~\citep{trove,genprove}. The schema associates evidence with the answer as a whole, with source granularity ranging from supporting sentences to necessary figure or table components.

\begin{figure}[t]
    \centering
    \includegraphics[width=\linewidth]{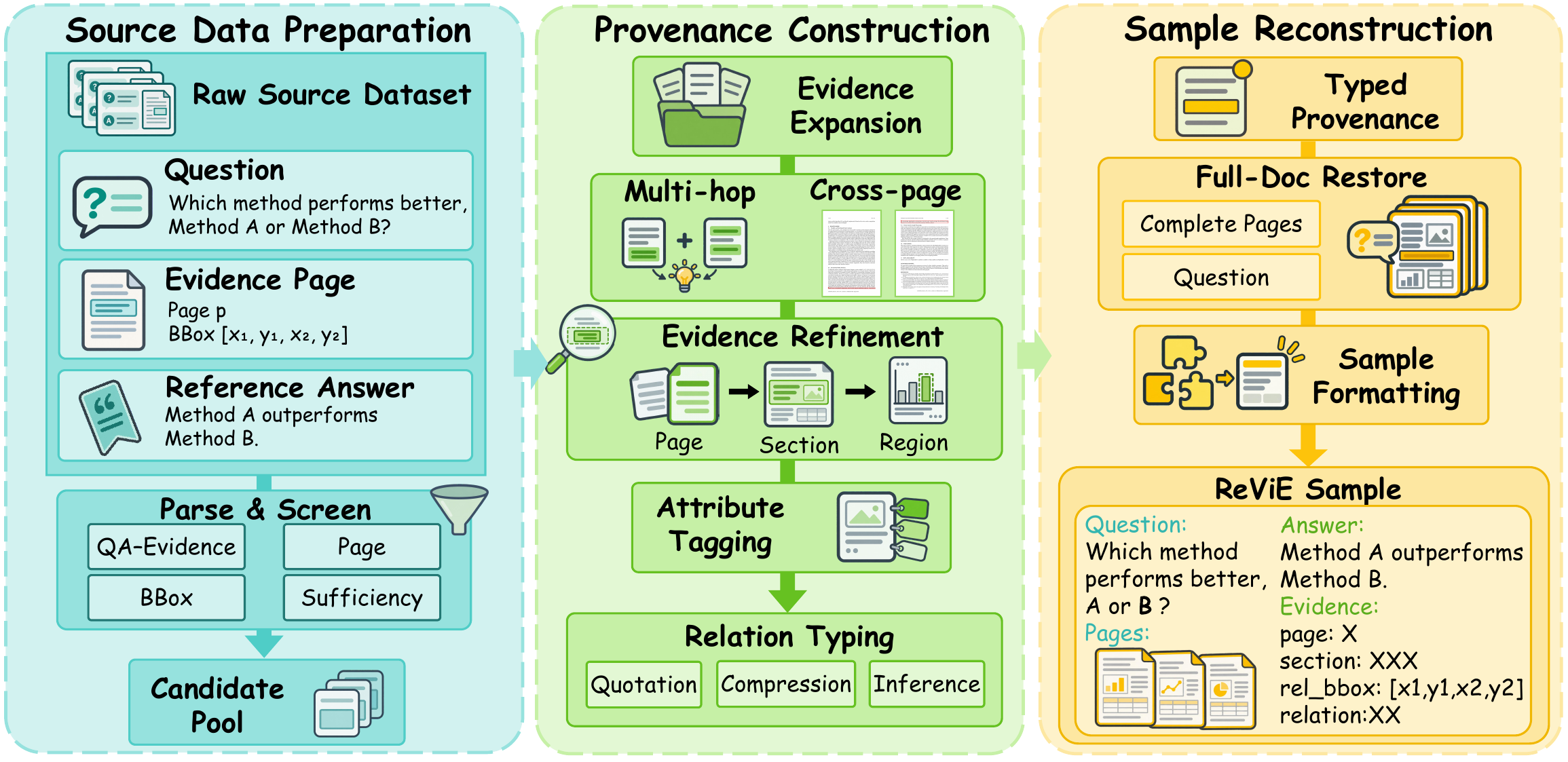}
    \caption{ReViE construction. Source QA pairs and newly generated, verified multi-hop QA pairs undergo fixed-QA evidence refinement. A later branch constructs additional QA pairs from contiguous evidence across adjacent pages. Supporting regions receive relation annotations, and complete ordered document pages are restored for every retained instance.}
    \label{fig:data_construction}
\end{figure}

\subsection{Dataset Construction}
\label{sec:dataset_construction}

ReViE combines source QA pairs from SciEGQA~\citep{sciegqa} with newly constructed multi-hop and cross-page-contiguous QA pairs (Figure~\ref{fig:data_construction}). GPT-5.6 Sol~\citep{openai2026gpt56} assists evidence refinement and relation annotation, complemented by human annotator review and second-round expert validation to ensure evidence quality and annotation consistency.

\textbf{QA preparation and multi-hop construction.}
We associate source annotations with page images and screen candidate QA--evidence associations for consistency, correct evidence pages, valid boxes, and sufficient support. In an independent image-direct stage, new questions and reference answers are generated from at least two visual evidence regions. A blind visual judge checks whether the complete answer genuinely requires combining multiple regions. Verified QA pairs are then fixed for subsequent refinement with their question--answer content preserved.

\textbf{Fixed-QA evidence refinement.}
This path processes both source QA pairs and previously constructed, verified multi-hop QA pairs. It preserves the question, reference answer, evidence-page identities, and correspondence to the original evidence anchors. Refinement selects complete supporting sentences or necessary figure and table components within the supplied evidence crops. Individual regions contribute support for particular answer content, while the evidence set must collectively support the complete reference answer.

\textbf{Cross-page-contiguous QA construction.}
A later branch constructs additional QA pairs from continuous supporting content across adjacent-page boundaries. Valid configurations include a sentence split across pages or semantically continuous adjacent sentences that jointly support the QA pair. New questions and reference answers are grounded in the boundary evidence without outside knowledge. Cross-page continuity does not imply verified multi-hop status.

\textbf{Annotation and finalization.}
Both later processing paths record the exact real section or subsection heading when it can be reliably established; otherwise, they use JSON \texttt{null}. Once a QA pair and its evidence are fixed, each region receives a support relation according to its contribution to the complete answer. Source units split across boxes are reconstructed before relation assignment, and uncertain cases enter a second-pass audit. The final quality-control pipeline retains 3,941 of 4,168 candidates after excluding 218 instances requiring box repair and nine with evidence mismatches. We restore complete ordered document pages and partition instances at the document level. Appendix~\ref{app:dataset_quality} details quality control, Appendix~\ref{app:refinement_prompt} describes the construction protocol, and Figure~\ref{fig:relation_annotation_prompt} presents the relation-annotation instructions.

\subsection{Dataset Analysis}
\label{sec:dataset_analysis}
\begin{wrapfigure}[19]{r}{0.47\textwidth}
    \vspace{-2.2em}
    \centering
    \includegraphics[width=\linewidth]{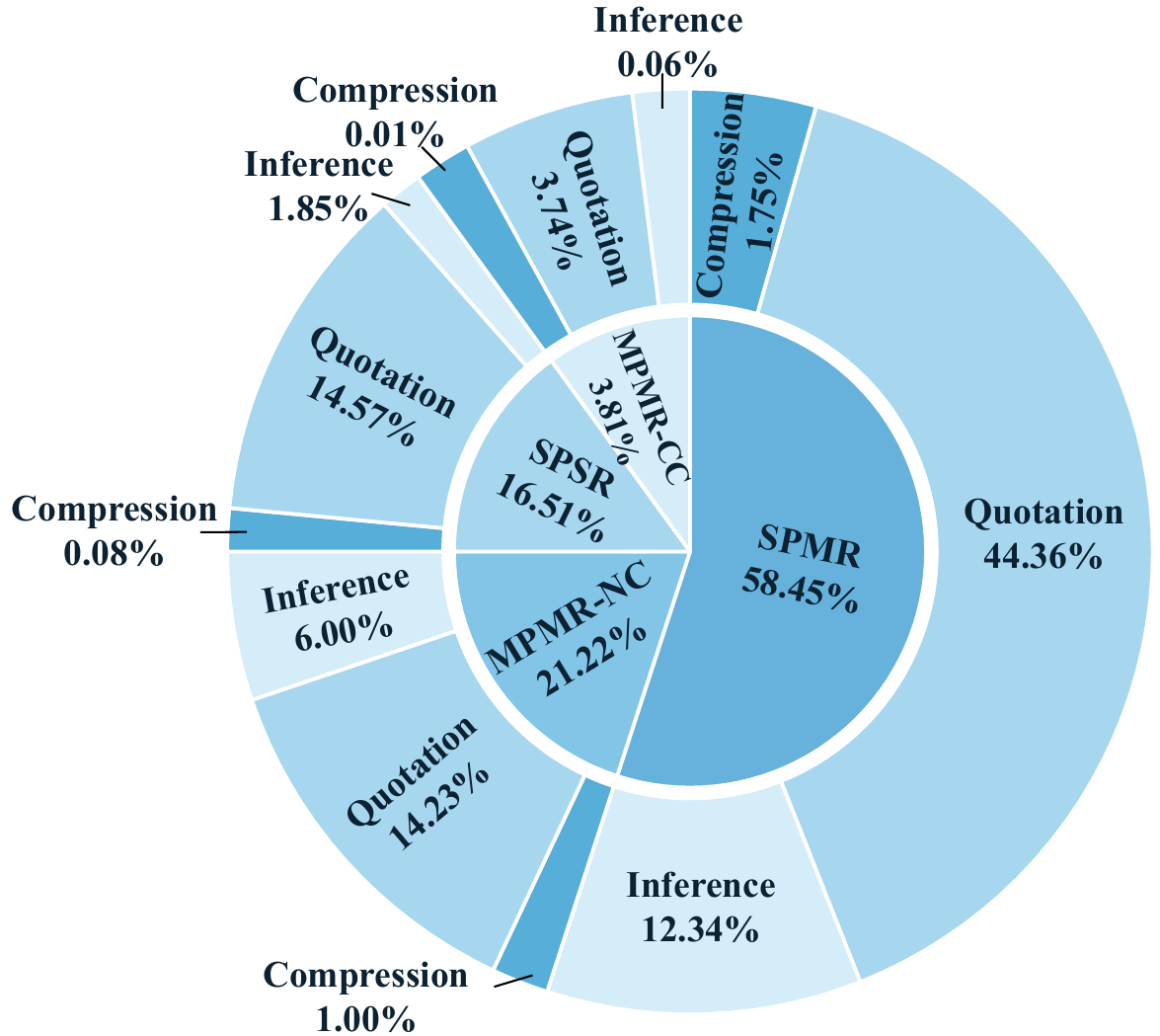}
    \caption{Evidence-region distribution by spatial structure and support relation. Percentages are over 8,523 annotated regions.}
    \label{fig:data_analysis}
\end{wrapfigure}

\textbf{Coverage and spatial structure.}
ReViE contains 3,941 instances, 1,612 documents, 31,981 unique page images, and 8,523 evidence regions across eight scientific domains. Its document-disjoint training, validation, and test splits contain 3,115, 417, and 409 instances. Single-page single-region (SPSR), single-page multi-region (SPMR), and multi-page multi-region (MPMR) instances account for 35.70\%, 46.18\%, and 18.12\%, respectively. Figures, tables, or mixed evidence occur in 28.88\% of instances.

\textbf{Support relations and sparsity.}
At the region level, \emph{Quotation}, \emph{Compression}, and \emph{Inference} account for 76.91\%, 2.84\%, and 20.25\%. Figure~\ref{fig:data_analysis} separates MPMR instances without and with a cross-page-contiguity annotation: the former has the highest within-group \emph{Inference} proportion, while the latter mainly contains \emph{Quotation}. An instance has 2.16 evidence regions on 1.18 pages on average, within an 18.28-page input document. The task therefore requires locating a small set of supporting regions within a much longer document. Additional statistics appear in Appendix~\ref{app:dataset_statistics}.

\subsection{Benchmark Comparison}
\label{sec:benchmark_comparison}

Existing benchmarks evaluate multi-page document QA or spatial evidence grounding~\citep{mmlongbenchdoc,sciegqa,m3grounder}. ReViE combines full scientific-document input with page, section, and region annotations and a semantic relation for each evidence item. It additionally records multi-region support, multi-page support, cross-page continuity, and multi-hop configurations. Table~\ref{tab:benchmark_comparison} compares the documented annotation schemes; these dimensions jointly describe task coverage and the granularity of available supervision.

\section{VisProve}
\label{sec:method}

\begin{figure}[t]
    \centering
    \includegraphics[width=\linewidth]{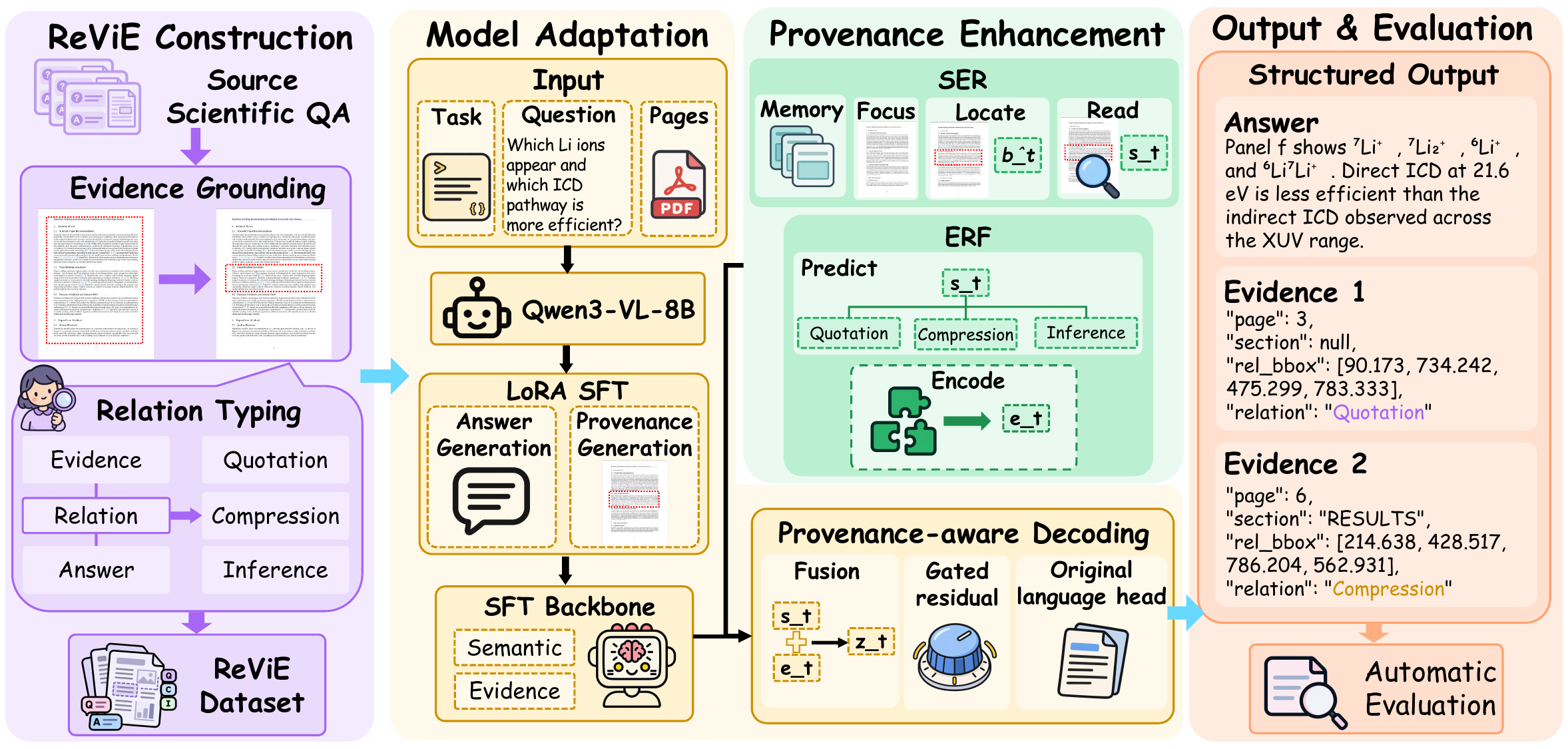}
    \caption{VisProve. Joint SFT learns answer--provenance generation. SER reads localized visual features, ERF encodes their support relation, and a gated residual combines the two representations before the original language head.}
    \label{fig:method_overview}
\end{figure}

VisProve learns joint answer--provenance generation and augments it with localized visual reading and relation feedback. Figure~\ref{fig:method_overview} summarizes the framework. SER forms a spatial evidence representation from document features, ERF encodes its support relation, and a gated residual conditions the language head on both signals.

\subsection{Joint Answer--Provenance Learning}
\label{sec:sft}

We fine-tune Qwen3-VL-8B on a unified target containing the complete answer and its structured evidence set. The standard causal language-modeling objective supervises the entire target sequence. This SFT model provides the common generation backbone; its parameters are then frozen while the provenance components are trained on the same data. Appendix~\ref{app:sft_configuration} specifies the SFT objective and configuration.

\subsection{Spatial Evidence Reading}
\label{sec:ser}

SER connects a latent location prediction to differentiable reading from the corresponding spatial feature grid. At generation step $t$, the decoder state $h_t$ depends on the document, question, and generated prefix.

For each page $j$, frozen visual features are pooled into a $16\times16$ grid and projected to a \(d=128\)-dimensional visual memory $\{v_{jk}\}$ with spatial positional encodings. The memory contains all document pages. Using $h_t$ and the mean question embedding $\bar e_q$, SER computes a query and soft page distribution:
\begin{equation}
    u_t
    =
    W_q\,\mathrm{LN}(h_t)
    +
    W_{\mathrm{question}}\bar e_q,
    \qquad
    \pi_{tj}
    =
    \mathrm{softmax}_j
    \left(
        \frac{u_t^{\top}\bar v_j}{\sqrt{d}}
    \right),
    \label{eq:ser_query_page}
\end{equation}
where $\bar v_j$ is the mean visual-memory representation of page $j$.

A box predictor maps $u_t$ to an ordered normalized box $\hat b_t$. Smooth boundary functions convert this box into spatial weights $\alpha_{tjk}$, which pool local evidence alongside page-level context:
\begin{equation}
\begin{aligned}
    r_t^{\mathrm{vis}}
    &=
    \sum_j \pi_{tj}
    \sum_k \alpha_{tjk}v_{jk},
    \qquad
    c_t
    =
    \sum_j \pi_{tj}\bar v_j,\\
    s_t
    &=
    \mathrm{GELU}
    \left(
        W_f[u_t;\,r_t^{\mathrm{vis}};\,c_t]
    \right)
    +
    \mathrm{MLP}_{\mathrm{geometry}}(\hat b_t).
\end{aligned}
\label{eq:ser_read}
\end{equation}

The spatial representation $s_t$ combines local appearance, page context, the question-conditioned query, and box geometry. The internal box guides feature reading, while the final output coordinates are separately generated by the language head. Appendix~\ref{app:ser_erf_training} defines the box parameterization and spatial weights.

\subsection{Evidence--Answer Relation Feedback}
\label{sec:erf}

Given the question-conditioned spatial representation $s_t$, ERF predicts a distribution over three support relations and maps it to a continuous feedback embedding for next-token prediction:
\begin{equation}
\begin{aligned}
    z_t^{\mathrm{rel}} = W_{\mathrm{rel}}s_t+b_{\mathrm{rel}},\quad
    \rho_t = \mathrm{softmax}(z_t^{\mathrm{rel}}),\quad
    e_t = W_{\mathrm{type}}\rho_t .
\end{aligned}
\label{eq:erf}
\end{equation}
The entries of $\rho_t\in\mathbb{R}^{3}$ correspond to \emph{Quotation}, \emph{Compression}, and \emph{Inference}. Since $s_t$ incorporates both visual evidence and $h_t$, the relation estimate is conditioned on the question and generated prefix. The embedding $e_t$ provides semantic feedback to generation; the final relation field is produced by the original language head.

\subsection{Joint Fusion and Training Objective}
\label{sec:joint_fusion}

\textbf{Gated residual fusion.}
The full model combines the spatial and relation representations as
\begin{equation}
    z_t=s_t+e_t.
    \label{eq:module_combination}
\end{equation}
It projects this representation to the decoder dimension and applies a bounded residual update:
\begin{equation}
    \Delta h_t=g_t\lambda_t W_{\mathrm{up}}z_t,
    \qquad h'_t=h_t+\Delta h_t.
    \label{eq:gated_residual}
\end{equation}
The learned gate $g_t$ scales the update, and $\lambda_t$ limits its pre-gated norm to $3\%$ of $\lVert h_t\rVert_2$. The update is applied after the final decoder layer and before the original output head, which predicts the next token from $h'_t$. Appendix~\ref{app:ser_erf_training} provides the gate and normalization formulas. Section~\ref{sec:experimental_setup} defines the feedback variants.

\textbf{Training objective.}
With the SFT backbone frozen, the added components are optimized using the structured generation objective together with auxiliary provenance supervision:
\begin{equation}
\begin{aligned}
    \mathcal{L}
    ={}&
    \mathcal{L}_{\mathrm{gen}}
    +
    \lambda_{\mathrm{loc}}\,
    \mathbb{I}_{\mathrm{SER}}
    \left(
        \mathcal{L}_{\mathrm{page}}
        +
        \mathcal{L}_{\mathrm{box}}
    \right)
    +
    \lambda_{\mathrm{rel}}\,
    \mathbb{I}_{\mathrm{ERF}}
    \mathcal{L}_{\mathrm{rel}}\\
    &+
    \lambda_{\mathrm{field}}
    \mathcal{L}_{\mathrm{field}}
    +
    \lambda_{\mathrm{outside}}
    \mathcal{L}_{\mathrm{outside}}
    +
    \lambda_{\mathrm{gate}}\,
    \mathbb{I}_{\mathrm{SER}}
    \mathcal{L}_{\mathrm{gate}}.
\end{aligned}
\label{eq:overall_objective}
\end{equation}
The indicators $\mathbb{I}_{\mathrm{SER}}$ and $\mathbb{I}_{\mathrm{ERF}}$ specify whether the corresponding components are optimized. The page and box losses supervise latent localization, the relation loss supervises the three-way relation prediction, and the remaining losses weight provenance fields and regularize residual feedback. Auxiliary supervision pairs serialized evidence items with their gold annotations in order and is applied at configuration-specific decoder positions. Gold annotations provide training targets; inference uses the question and document pages. Appendix~\ref{app:ser_erf_training} specifies the supervision positions, trainable components, loss reductions, and weights.

\section{Experiments}
\label{sec:experiments}

\subsection{Experimental Setup}
\label{sec:experimental_setup}

\textbf{Data and models.}
We train on the 3,115-instance ReViE training split and evaluate on the document-disjoint 409-instance test split. Models receive the question and complete ordered document pages. General-purpose baselines comprise Qwen3-VL (8B and 32B)~\citep{qwen3vl}, InternVL3.5 (8B and 38B)~\citep{internvl35}, MiniCPM-V4.5-8B~\citep{minicpmv45}, Qwen2.5-VL-7B~\citep{qwen25vl}, Eagle2.5-8B~\citep{eagle25}, GPT-4o-mini~\citep{gpt4omini}, Claude Haiku 4.5~\citep{claudehaiku45}, and Gemini 2.5 Flash-Lite~\citep{gemini25flashlite}. These models and the VisProve variants use the prompt in Appendix~\ref{app:inference_prompt}.

\textbf{Training and feedback variants.}
VisProve uses LoRA-based SFT~\citep{lora} of Qwen3-VL-8B followed by provenance-component training with the SFT backbone frozen. The base model is included in the main comparison, while Default SFT provides the common backbone for the ablation variants. \emph{w/o ERF} uses spatial feedback only ($z_t=s_t$) and disables ERF. \emph{w/o SER} uses relation feedback only ($z_t=e_t$), retaining a frozen spatial reader to supply ERF's input but excluding $s_t$ from direct residual feedback. The full model jointly optimizes and injects both representations. Auxiliary field losses follow the active components, as specified in Appendix~\ref{app:ser_erf_training}. All configurations generate their final fields through the language head, without post-hoc prediction repair.

\textbf{Metrics.}
Answer quality is measured by ROUGE-L~\citep{rouge} and BERTScore-F1~\citep{bertscore}; localization by Page-F1, Section-F1, Evidence-IoU, and Region-F1; and typed provenance by Relation-F1 and Typed-F1. All are reported on a 0--100 scale. Typed-F1 evaluates all test instances using page-, section-, and relation-consistent region matching. Evidence-IoU, Region-F1, and Relation-F1 are conditional averages over each model's instances with an exactly correct evidence-page set; Section-F1 uses instances with a non-null reference section. Overall averages the answer, localization, and typed-provenance category scores with equal weight. A text-only judge scores answer quality on 50 shared questions using a 0--5 scale; its mean score is reported separately as Judge. Appendix~\ref{app:evaluation} defines matching and aggregation.

\subsection{Main Results and Ablation Study}
\label{sec:main_results}
\begin{table}[t]
\vspace{-1.0em}
\centering
\caption{Main results on ReViE. E-IoU, R-F1, and Rel-F1 condition on exact page-set prediction; T-F1 uses all 409 test instances. Overall equally averages the three metric categories. Judge uses 50 shared questions. Light orange and light green mark the best and second-best values, respectively.}
\label{tab:main_results}

\footnotesize
\setlength{\tabcolsep}{2.8pt}
\renewcommand{\arraystretch}{1.10}

\begin{tabular}{l|cc|cccc|cc|cc}
\toprule

\textbf{Model}
& \multicolumn{2}{c|}{\textbf{Answer}}
& \multicolumn{4}{c|}{\textbf{Localization}}
& \multicolumn{2}{c|}{\textbf{Typed Prov.}}
& \multicolumn{2}{c}{\textbf{Summary}} \\

& \textbf{R-L$\uparrow$}
& \textbf{B-F1$\uparrow$}
& \textbf{P-F1$\uparrow$}
& \textbf{S-F1$\uparrow$}
& \textbf{E-IoU$\uparrow$}
& \textbf{R-F1$\uparrow$}
& \textbf{Rel-F1$\uparrow$}
& \textbf{T-F1$\uparrow$}
& \textbf{Overall$\uparrow$}
& \textbf{Judge$\uparrow$} \\

\midrule

\multicolumn{11}{l}{\textbf{Open-source models}} \\

Qwen3-VL-8B
& \second{\underline{48.27}}
& 86.08
& 66.55
& 31.90
& 1.32
& 1.60
& 0.73
& 0.18
& \second{\underline{30.99}}
& 3.04 \\

InternVL3.5-8B
& 26.63
& 84.26
& 21.81
& 14.35
& 0.54
& 0.76
& 0.66
& 0.00
& 21.71
& 2.66 \\

MiniCPM-V4.5-8B
& 10.73
& 80.99
& 1.88
& 4.63
& 0.00
& 0.00
& 0.00
& 0.00
& 15.83
& 1.88 \\

Qwen3-VL-32B
& 33.24
& 84.70
& 66.16
& \second{\underline{37.01}}
& 1.77
& 2.20
& 0.93
& 0.87
& 28.89
& 3.34 \\

InternVL3.5-38B
& 4.98
& 79.91
& 4.38
& 3.77
& 1.48
& 2.36
& 2.36
& 0.00
& 15.54
& 2.64 \\

Qwen2.5-VL-7B
& 44.39
& \second{\underline{88.09}}
& 32.29
& 12.04
& 0.02
& 0.04
& 0.00
& 0.04
& 25.79
& 2.44 \\

Eagle2.5-8B
& 35.86
& 85.69
& 28.13
& 13.89
& 0.97
& 1.36
& 1.30
& 0.08
& 24.18
& 2.24 \\

\midrule

\multicolumn{11}{l}{\textbf{Closed-source models}} \\

GPT-4o-mini
& 18.88
& 85.67
& 31.46
& 14.35
& 0.45
& 0.59
& 0.13
& 0.23
& 21.39
& 1.36 \\

Claude Haiku 4.5
& 15.76
& 84.86
& 36.35
& 11.54
& 2.34
& 2.73
& 0.97
& 0.17
& 21.37
& 1.48 \\

Gemini 2.5 Flash-Lite
& 23.41
& 84.52
& \second{\underline{73.81}}
& 32.10
& \second{\underline{6.59}}
& \second{\underline{7.08}}
& \second{\underline{6.68}}
& \second{\underline{3.21}}
& 29.60
& \second{\underline{3.42}} \\

\midrule

\textbf{VisProve (Ours)}
& \best{67.08}
& \best{94.18}
& \best{75.73}
& \best{53.24}
& \best{8.58}
& \best{9.68}
& \best{9.22}
& \best{5.48}
& \best{41.60}
& \best{3.46} \\

\bottomrule
\end{tabular}
\end{table}

Table~\ref{tab:main_results} compares VisProve with general-purpose open- and closed-source VLMs, while Table~\ref{tab:ablation_results} reports the comparison with Default SFT and the two feedback variants.

\begin{table}[t]
\centering
\caption{Ablation study on ReViE. Metric definitions, evaluation conditions, and best/second-best highlighting follow Table~\ref{tab:main_results}.}
\label{tab:ablation_results}

\footnotesize
\setlength{\tabcolsep}{2.8pt}
\renewcommand{\arraystretch}{1.10}

\begin{tabular}{l|cc|cccc|cc|cc}
\toprule

\textbf{Model}
& \multicolumn{2}{c|}{\textbf{Answer}}
& \multicolumn{4}{c|}{\textbf{Localization}}
& \multicolumn{2}{c|}{\textbf{Typed Prov.}}
& \multicolumn{2}{c}{\textbf{Summary}} \\

& \textbf{R-L$\uparrow$}
& \textbf{B-F1$\uparrow$}
& \textbf{P-F1$\uparrow$}
& \textbf{S-F1$\uparrow$}
& \textbf{E-IoU$\uparrow$}
& \textbf{R-F1$\uparrow$}
& \textbf{Rel-F1$\uparrow$}
& \textbf{T-F1$\uparrow$}
& \textbf{Overall$\uparrow$}
& \textbf{Judge$\uparrow$} \\

\midrule

\textbf{VisProve (Ours)}
& \second{\underline{67.08}}
& \best{94.18}
& \best{75.73}
& \best{53.24}
& \best{8.58}
& \best{9.68}
& \best{9.22}
& \best{5.48}
& \best{41.60}
& 3.46 \\

w/o ERF
& 66.85
& 93.92
& \second{\underline{75.21}}
& 45.06
& 6.66
& 7.60
& 7.03
& 3.73
& 39.80
& \best{3.56} \\

w/o SER
& \best{67.82}
& \second{\underline{93.96}}
& 74.69
& \second{\underline{48.61}}
& \second{\underline{6.86}}
& \second{\underline{7.94}}
& \second{\underline{7.29}}
& \second{\underline{4.11}}
& \second{\underline{40.37}}
& \second{\underline{3.54}} \\

Default SFT
& 64.24
& 93.85
& 75.00
& 45.99
& 5.07
& 5.91
& 5.22
& 3.52
& 38.80
& 3.52 \\

\bottomrule
\end{tabular}
\end{table}

\textbf{Task adaptation and typed provenance.}
Tables~\ref{tab:main_results} and~\ref{tab:ablation_results} show gains from task-specific SFT and provenance feedback. Relative to the Qwen3-VL-8B base, Default SFT raises ROUGE-L from 48.27 to 64.24 and Typed-F1 from 0.18 to 3.52. VisProve further improves Typed-F1 to 5.48 and Section-F1 from 45.99 to 53.24, while Page-F1 increases only modestly from 75.00 to 75.73, indicating larger gains in finer provenance outputs than in page identification. In the main comparison, VisProve achieves the highest Overall score of 41.60 and leads all eight task-specific metrics.

\textbf{Feedback configurations.}
Table~\ref{tab:ablation_results} compares the feedback variants on a shared SFT backbone. Spatial feedback only (\emph{w/o ERF}) reaches 3.73 Typed-F1, while relation feedback with a frozen spatial reader (\emph{w/o SER}) reaches 4.11; combining both representations yields 5.48. VisProve also raises Overall from 38.80 for Default SFT to 41.60 and leads seven of the eight task-specific metrics, although \emph{w/o SER} achieves the highest ROUGE-L and \emph{w/o ERF} the highest Judge score. These results support the full configuration's advantage in typed provenance without implying uniform gains across answer-quality measures; the variants differ in feedback and associated training configurations rather than the SFT backbone.

\subsection{Comparison with Existing Methods}
\label{sec:method_comparison}

Table~\ref{tab:method_comparison} compares VisProve with VISA~\citep{visa}, Chain of Evidence (CoE)~\citep{chainofevidence}, and UniDoc-RL-7B~\citep{unidocrl} in their shared answer--page--region output space. Their native inference procedures are evaluated with ROUGE-L, BERTScore-F1, Page-F1, Evidence-IoU, and Region-F1. The comparison-specific score $\mathrm{Overall}_{\mathrm{shared}}$ equally weights answer quality and localization (Appendix~\ref{app:overall_score}).

\begin{table}[h]
\vspace{-1.0em}
\centering
\caption{Comparison in the shared answer--page--region output space. $\mathrm{Overall}_{\mathrm{shared}}$ uses two categories and differs from Overall in Table~\ref{tab:main_results}. E-IoU and R-F1 retain the page-exact conditioning. Light orange and light green mark the best and second-best values, respectively.}
\label{tab:method_comparison}

\small
\setlength{\tabcolsep}{5.0pt}
\renewcommand{\arraystretch}{1.12}

\begin{tabular}{l|cc|ccc|c}
\toprule
\textbf{Method}
& \multicolumn{2}{c|}{\textbf{Answer}}
& \multicolumn{3}{c|}{\textbf{Localization}}
& \textbf{Summary} \\

& \textbf{R-L$\uparrow$}
& \textbf{B-F1$\uparrow$}
& \textbf{P-F1$\uparrow$}
& \textbf{E-IoU$\uparrow$}
& \textbf{R-F1$\uparrow$}
& \textbf{Overall$_{\mathrm{shared}}\!\uparrow$} \\
\midrule

VISA~\citep{visa}
& 11.65 & 85.53 & 7.50 & 0.02 & 0.02 & 25.55 \\

CoE~\citep{chainofevidence}
& 10.45 & 84.57 & 30.42
& \second{\underline{1.37}}
& \second{\underline{1.55}}
& 29.31 \\

UniDoc-RL-7B~\citep{unidocrl}
& \second{\underline{28.37}}
& \second{\underline{87.10}}
& \second{\underline{65.31}}
& 0.60
& 0.77
& \second{\underline{39.98}} \\

\midrule

\textbf{VisProve (Ours)}
& \best{67.08}
& \best{94.18}
& \best{75.73}
& \best{8.58}
& \best{9.68}
& \best{55.98} \\

\bottomrule
\end{tabular}
\end{table}

\textbf{Results.}
VisProve has the highest values on all five shared metrics and reaches 55.98 $\mathrm{Overall}_{\mathrm{shared}}$. Among the baselines, UniDoc-RL achieves the highest answer-quality and page-level scores, while CoE obtains higher conditional region-overlap scores (Table~\ref{tab:method_comparison}). This comparison measures VisProve after ReViE-specific training against the existing methods under their native inference procedures; the shared-backbone comparison with Default SFT provides the task-trained reference point.

\subsection{Qualitative Error Analysis}
\label{sec:qualitative_error_analysis}

The examples in Appendix~\ref{app:qualitative_error_analysis} show several discrepancies between answers and provenance. One prediction gives a correct chart-reading answer but cites adjacent prose. Another omits a supporting page and misidentifies two ion species. A third omits a legend region, gives an underspecified answer, and predicts a relation different from the annotation. These cases show that answer correctness does not necessarily imply accurate visual provenance, and that localization, evidence coverage, and relation typing can fail jointly. A mismatch with the annotated evidence also does not necessarily exclude other valid supporting regions. These observations motivate evaluating source location, evidence coverage, and semantic support alongside answer quality.
\begin{wrapfigure}{R}{0.47\textwidth}
    \vspace{-2.0em}
    \centering
    \includegraphics[width=\linewidth]{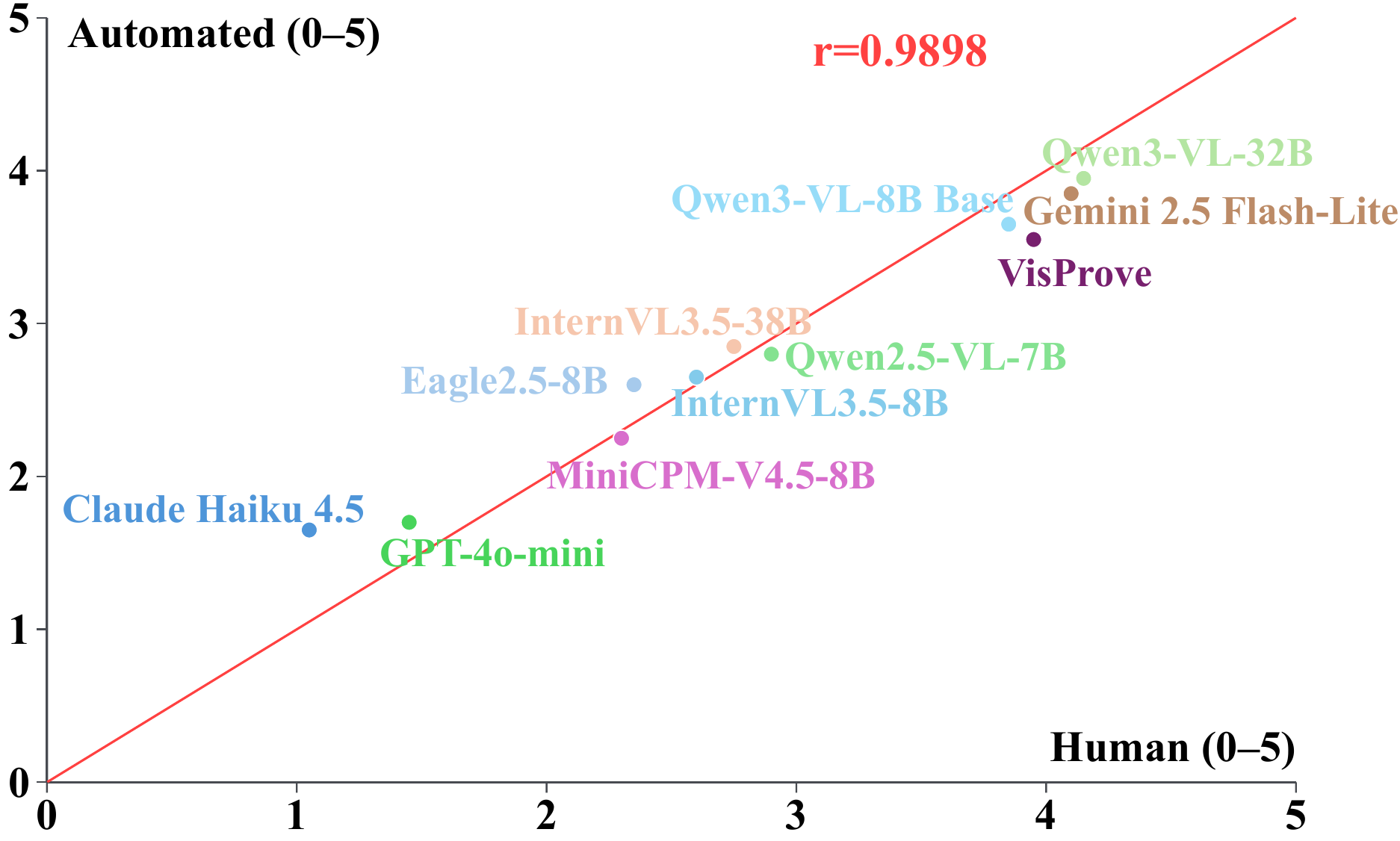}
    \caption{Consistency between LLM-as-Judge scores and human ratings across 11 evaluated models.}
    \label{fig:human_consistency}
    \vspace{-0.6em}
\end{wrapfigure}
\subsection{Consistency with Human Judgments}
\label{sec:human_consistency}

Human reviewers assess 220 answers from 11 models on the same 20 test questions, using the question, reference answer, and predicted answer. Reviewers score answers independently without access to model identities or automated scores, and the final human rating for each answer is the median of its individual ratings. The model-level mean ratings have Pearson correlation $r=0.9898$ with Judge scores on those same questions (Figure~\ref{fig:human_consistency}). This analysis concerns agreement across 11 model averages in reference-based answer assessment. Appendix~\ref{app:human_evaluation} describes the independent ratings and aggregation procedure.

\section{Conclusion}
\label{sec:conclusion}

Visual provenance should reveal both where supporting evidence appears and how it contributes to an answer. ReViE makes source localization and support relations jointly evaluable, while VisProve couples differentiable spatial evidence reading with support relation feedback during generation. Using the same backbone and training split, VisProve improves typed provenance and section-level localization over joint SFT, although gains in answer quality are not uniform. Meanwhile, the low absolute typed provenance scores highlight a substantial gap between producing a correct answer and precisely attributing its visual support. These results support learning evidence localization and evidence use jointly, while treating provenance as an evaluation target distinct from answer quality.


\subsection*{AI Use Statement}
Generative AI tools were used to assist dataset construction, including QA construction, visual-evidence refinement, multi-hop verification, and evidence--answer relation annotation. The authors take responsibility for the final dataset, reported results, and conclusions.

\subsection*{Reproducibility Statement}
Appendix~\ref{app:dataset_details} documents dataset construction, statistics, and quality-control procedures; Appendix~\ref{app:implementation} provides training and inference configurations; and Appendix~\ref{app:evaluation} specifies metric definitions and automated and human evaluation protocols.

\bibliography{iclr2027_conference}
\bibliographystyle{iclr2027_conference}

\newpage
\appendix
\section{Additional Dataset Details}
\label{app:dataset_details}

\subsection{Additional Dataset Statistics}
\label{app:dataset_statistics}

ReViE contains 3,941 instances from 1,612 scientific documents, comprising 31,981 unique page images and 8,523 annotated evidence regions. Table~\ref{tab:dataset_split_statistics} reports the split-level statistics. Splits are constructed at the document level, with no document or sample overlap across training, validation, and test sets.

\begin{table}[htbp]
    \centering
    \caption{Split-level statistics of ReViE. Page counts refer to unique document pages within each split.}
    \label{tab:dataset_split_statistics}
    \small
    \setlength{\tabcolsep}{7pt}
    \renewcommand{\arraystretch}{1.12}
    \begin{tabular}{lrrrr}
        \toprule
        \textbf{Split}
        & \textbf{Instances}
        & \textbf{Documents}
        & \textbf{Pages}
        & \textbf{Evidence regions} \\
        \midrule
        Train      & 3,115 & 1,296 & 25,439 & 6,738 \\
        Validation &   417 &   161 &  3,106 &   921 \\
        Test       &   409 &   155 &  3,436 &   864 \\
        \midrule
        \textbf{Total}
        & \textbf{3,941}
        & \textbf{1,612}
        & \textbf{31,981}
        & \textbf{8,523} \\
        \bottomrule
    \end{tabular}
\end{table}

\textbf{Domain distribution.}
ReViE covers eight scientific domains. Table~\ref{tab:domain_distribution} reports both instance- and document-level distributions, showing broad coverage across computer science, economics, engineering, mathematics, physics, quantitative biology, quantitative finance, and statistics.

\begin{table}[htbp]
    \centering
    \caption{Domain distribution of ReViE at the instance and document levels.}
    \label{tab:domain_distribution}
    \small
    \setlength{\tabcolsep}{6pt}
    \renewcommand{\arraystretch}{1.10}
    \begin{tabular}{lrrrr}
        \toprule
        \textbf{Domain}
        & \textbf{Instances}
        & \textbf{Instance \%}
        & \textbf{Documents}
        & \textbf{Document \%} \\
        \midrule
        Computer Science
        & 493 & 12.51 & 200 & 12.41 \\

        Economics
        & 330 & 8.37 & 169 & 10.48 \\

        Electrical Engineering and Systems Science
        & 662 & 16.80 & 232 & 14.39 \\

        Mathematics
        & 107 & 2.72 & 59 & 3.66 \\

        Physics
        & 480 & 12.18 & 184 & 11.41 \\

        Quantitative Biology
        & 555 & 14.08 & 273 & 16.94 \\

        Quantitative Finance
        & 940 & 23.85 & 345 & 21.40 \\

        Statistics
        & 374 & 9.49 & 150 & 9.31 \\

        \midrule
        \textbf{Total}
        & \textbf{3,941}
        & \textbf{100.00}
        & \textbf{1,612}
        & \textbf{100.00} \\
        \bottomrule
    \end{tabular}
\end{table}

\textbf{Spatial and reasoning structure.}
At the instance level, ReViE contains 1,407 SPSR instances (35.70\%), 1,820 SPMR instances (46.18\%), and 714 MPMR instances (18.12\%). Among all instances, 604 (15.33\%) are labeled as multi-hop and 162 (4.11\%) as cross-page contiguous. These attributes describe different properties: multi-hop concerns the necessity of combining evidence regions, whereas cross-page continuity concerns supporting content across an adjacent-page boundary. The two labeled subsets do not overlap in the final dataset. This dataset-specific observation does not imply conceptual incompatibility: multi-hop evidence may span multiple pages without being labeled as cross-page contiguous.

\textbf{Evidence modality, granularity, and relation distribution.}
Table~\ref{tab:modality_relation_distribution} summarizes evidence modality at the instance level and evidence--answer relations at the evidence-region level. ReViE is predominantly text-based, while figures, tables, and mixed-modality evidence together account for 28.88\% of instances. At the evidence-granularity level, 2,803 instances use sentence-level evidence, 836 use localized visual regions, 24 require complete visual regions, and 278 combine multiple granularities. \emph{Quotation} is the dominant relation type, while \emph{Inference} forms a substantial minority and \emph{Compression} is comparatively rare. A total of 482 instances (12.23\%) contain more than one evidence--answer relation type.

\textbf{Evidence density and document length.}
Each instance contains 2.16 evidence regions and 1.18 unique evidence pages on average. The complete document input contains 18.28 pages on average (median 14), with a maximum of 234 pages. The median number of evidence regions per instance is 2, and 90\% of instances contain at most four evidence regions. No reliable section heading can be identified for 2,383 evidence regions (27.96\%); these cases use JSON \texttt{null} in the final annotation schema.

\begin{table}[htbp]
    \centering
    \caption{Evidence modality and evidence--answer relation distributions in ReViE. Modality statistics are computed over 3,941 instances, whereas relation statistics are computed over 8,523 evidence regions.}
    \label{tab:modality_relation_distribution}
    \small
    \setlength{\tabcolsep}{5pt}
    \renewcommand{\arraystretch}{1.10}
    \begin{tabular}{lrr@{\qquad}lrr}
        \toprule
        \multicolumn{3}{c}{\textbf{Evidence modality}}
        & \multicolumn{3}{c}{\textbf{Evidence--answer relation}} \\
        \cmidrule(lr){1-3}
        \cmidrule(lr){4-6}

        \textbf{Modality}
        & \textbf{Instances}
        & \textbf{\%}
        & \textbf{Relation}
        & \textbf{Regions}
        & \textbf{\%} \\
        \midrule

        Text
        & 2,803 & 71.12
        & Quotation
        & 6,555 & 76.91 \\

        Figure
        & 632 & 16.04
        & Compression
        & 242 & 2.84 \\

        Table
        & 212 & 5.38
        & Inference
        & 1,726 & 20.25 \\

        Mixed
        & 294 & 7.46
        & & & \\

        \midrule
        \textbf{Total}
        & \textbf{3,941}
        & \textbf{100.00}
        & \textbf{Total}
        & \textbf{8,523}
        & \textbf{100.00} \\
        \bottomrule
    \end{tabular}
\end{table}

\subsection{Quality Control and Split Integrity}
\label{app:dataset_quality}

This section describes qualification, verification, and review procedures. The construction operations and annotation conventions are detailed in Appendix~\ref{app:refinement_prompt}.

\textbf{Task-specific qualification.}
Candidate QA--evidence associations are screened for question--answer--evidence consistency, evidence-page correctness, bounding-box validity, and evidence sufficiency. Instances that do not provide reliable visual supervision are excluded from subsequent construction.

\textbf{Visual and structural verification.}
An independent visual quality check examines answer support, evidence sufficiency, localization precision, section annotations, and structural metadata. For multi-hop candidates, a blind visual judge checks whether the complete answer genuinely requires combining information from multiple evidence regions. This construction-stage verification is distinct from the text-only answer evaluation in Appendix~\ref{app:judge_protocol}. Cross-page-contiguous candidates are accepted only when the adjacent-page boundary content provides genuine continuous support.

\textbf{Relation review and final exclusions.}
After the QA pair and its visual evidence are fixed, first-pass relation annotation may return \texttt{NEEDS\_REVIEW} when the supplied evidence is unsupported or the relation is genuinely uncertain. Such cases undergo a separate second-pass audit, which may accept the existing evidence, flag a bounding-box repair requirement or evidence mismatch, or defer the instance for human review. The first-pass relation instructions are shown in Figure~\ref{fig:relation_annotation_prompt}. Across the final quality-control pipeline, 227 of 4,168 candidate instances are excluded, including 218 requiring bounding-box repair and nine with evidence mismatches, yielding the final 3,941-instance dataset.

\textbf{Human review and split integrity.}
The model-assisted construction process is complemented by human annotator review and second-round expert validation. We partition ReViE at the document level so that all instances from the same source document remain within a single split. The training, validation, and test sets contain 3,115, 417, and 409 instances, respectively, with no document or sample overlap across splits. Complete document pages are restored in their original order for every retained instance, and gold evidence pages are not used to preselect model inputs.

\subsection{QA and Visual Evidence Construction Protocol}
\label{app:refinement_prompt}

\textbf{Image-direct multi-hop construction and QA freezing.}
An independent image-direct stage constructs new questions and reference answers from at least two visual evidence regions. After passing the multi-hop verification described in Appendix~\ref{app:dataset_quality}, these QA pairs enter fixed-QA evidence refinement together with source QA pairs. Questions and reference answers are frozen for both groups during refinement.

\textbf{Anchor-preserving evidence refinement.}
Here, ``existing QA'' refers to a QA pair established before refinement, not necessarily inherited from the source dataset. Refinement preserves evidence-page identities and correspondence to the original evidence anchors. Each anchor defines a supplied evidence crop within which one or more refined regions are selected. The returned region boundaries may change, but the source anchor is not replaced and the question is not re-answered. Individual regions support the answer content they contribute, while the complete evidence set must collectively support the full reference answer.

\textbf{Later cross-page-contiguous QA construction.}
A separate later branch constructs additional QA pairs from supporting content at adjacent page boundaries, without outside knowledge. It covers a sentence split across pages or semantically continuous adjacent sentences that jointly support the QA pair. In \texttt{NEW\_CROSS\_PAGE\_QA} mode, an accepted instance has \texttt{cross\_page\_continuity\_valid} set to \texttt{true}. Candidate preparation sets \texttt{verified\_multihop} to \texttt{false} for this branch, so continuity alone does not confer a verified multi-hop label. This setting does not, by itself, establish that combining multiple regions is unnecessary.

\textbf{Evidence completeness.}
Textual refinement requests complete supporting sentences. For figures and tables, it selects necessary cells, rows, headers, panels, legends, or other local components, retaining complete visual regions when required for sufficient support. When a supporting sentence spans a page boundary, the selected regions jointly cover the complete sentence across both pages; each page-local fragment need not independently form a complete sentence.

\textbf{Section annotation.}
Both fixed-QA refinement and cross-page construction use the exact real section or subsection heading when it can be reliably established from the supplied visual context; otherwise, they use JSON \texttt{null}. An unresolved heading does not by itself invalidate an instance, and headings must not be fabricated to avoid a \texttt{null} value. In fixed-QA refinement, any supplied low-resolution page context is used to identify section headings, not to determine evidence coordinates.

\textbf{Intermediate localization representations.}
Fixed-QA refinement returns \texttt{crop\_bboxes} on a 0--1000 coordinate scale relative to each supplied evidence crop. The cross-page branch selects image-derived line identifiers for textual evidence, from which code computes bounding boxes, and uses page-relative coordinates for figure and table regions. These intermediate fields are distinct from the final \texttt{rel\_bbox} field, which always uses a 0--1000 scale relative to the complete document page.

\subsection{Relation Annotation Prompt}
\label{app:dataset_prompt}

Figure~\ref{fig:relation_annotation_prompt} presents the first-pass evidence--answer relation instructions applied after QA construction and visual evidence refinement. At this stage, the question, complete reference answer, and supporting regions are fixed regardless of whether the QA pair was inherited from the source dataset or newly constructed. GPT-5.6 Sol assigns one of \emph{Quotation}, \emph{Compression}, or \emph{Inference} to each evidence item. Cases marked as \texttt{NEEDS\_REVIEW} follow the second-pass audit described in Appendix~\ref{app:dataset_quality}.

The displayed output-format block specifies alternative field values and value ranges; it is a schema description rather than a literal JSON response. An actual response selects one allowed value for each field.

\begin{promptfigure}{First-pass Relation Annotation Prompt}
\small

You annotate semantic provenance relations for an \textbf{EXISTING, already verified scientific visual-document QA sample}. The Question, complete Reference Answer, and every supplied Evidence region are trusted and \textbf{FROZEN}. Do not rewrite the answer, change evidence, add evidence, delete evidence, or assess bbox geometry.

For \textbf{EACH} supplied evidence image E1, E2, ..., decide how the visual content in that evidence region participates in supporting the complete Reference Answer. The red rectangle marks the exact evidence region; surrounding pixels are only small reading context.

\textbf{Choose exactly one relation for each evidence region:}

\begin{itemize}
    \setlength{\itemsep}{0pt}
    \setlength{\parskip}{0pt}
    \item \textbf{Quotation:} The portion of the Answer supported by this evidence can be read directly from visible text, a table cell/value, a label, or an explicit visual statement, with at most minor grammatical or wording changes.
    \item \textbf{Compression:} The supported Answer content faithfully condenses, summarizes, or reorganizes information explicitly present in this evidence (possibly together with other evidence), without introducing a new conclusion.
    \item \textbf{Inference:} The supported Answer content is not explicitly stated in this evidence and requires comparison, calculation, logical deduction, visual trend interpretation, or joint reasoning with other evidence regions.
\end{itemize}

\textbf{Important rules:}

\begin{enumerate}
    \setlength{\itemsep}{0pt}
    \setlength{\parskip}{0pt}
    \item Treat the complete QA pair as one sample. Do \textbf{NOT} split or rewrite it.
    \item An evidence region need not support every clause in the complete Answer. Judge its relation to the particular Answer content that it actually supports.
    \item If multiple evidence boxes are fragments of one continuous source unit because a sentence, table row, or visual element is split by a line, column, or page boundary, reconstruct that unit first and assign the \textbf{SAME} relation to every fragment. A directly stated sentence split across boxes remains Quotation; the physical split alone does not make it Compression or Inference.
    \item If a conclusion can only be obtained by combining multiple evidence regions, label every region essential to that conclusion as Inference.
    \item Multi-hop is not automatically Inference. Combining explicitly stated facts without deriving a new conclusion can be Compression.
    \item If one region participates in more than one way, choose the relation required for the most substantive Answer content it supports, using the priority Inference $>$ Compression $>$ Quotation, and explain the mixed case.
    \item If a supplied region does not actually support any part of the Answer, or the relation is genuinely indeterminate, return status \texttt{NEEDS\_REVIEW}.
\end{enumerate}

\textbf{Output format.}

Return JSON only with this exact top-level structure:

{\ttfamily\footnotesize
\{
\quad "status": "VALID" or "NEEDS\_REVIEW",\\
\quad "overall\_reason": "brief sample-level assessment",\\
\quad "relations": [\\
\qquad \{\\
\qquad\quad "evidence\_index": 1,\\
\qquad\quad "relation": "Quotation" or "Compression" or "Inference",\\
\qquad\quad "supported\_answer\_content": "the exact Answer content supported",\\
\qquad\quad "justification": "brief evidence-based reason",\\
\qquad\quad "confidence": 0.0 to 1.0\\
\qquad \}\\
\quad ]\\
\}
}

Return exactly one \texttt{relations} item for every supplied evidence image, in index order. Do not include markdown fences or explanations outside the JSON object.

\end{promptfigure}

\captionof{figure}{Prompt used for first-pass evidence--answer relation annotation with GPT-5.6 Sol. The prompt assigns a typed provenance relation to each verified visual evidence region before subsequent review and validation.}
\label{fig:relation_annotation_prompt}

\subsection{Benchmark Comparison Details}
\label{app:benchmark_comparison}

Table~\ref{tab:benchmark_comparison} compares ReViE with DocVQA~\citep{docvqa}, MP-DocVQA~\citep{mpdocvqa}, SlideVQA~\citep{slidevqa}, MMLongBench-Doc~\citep{mmlongbenchdoc}, XL-DocBench~\citep{xldocbench}, GroundingDocQA-Bench~\citep{m3grounder}, and SciEGQA~\citep{sciegqa}. We compare documented input settings and answer-linked evidence annotations across eight dimensions. GroundingDocQA-Bench is abbreviated as GDQA-Bench and is not conflated with the larger GroundingDocQA training resource.

\textbf{Input and spatial criteria.}
Input distinguishes a single page (1), a capped multi-page context (Cap), and a full-document setting (Full). MP-DocVQA uses up to 20 pages, and SlideVQA retains the first 20 slides~\citep{mpdocvqa,slidevqa}. SciEGQA offers a full-document setting alongside evidence-page and evidence-crop settings~\citep{sciegqa}. MR requires multiple answer-linked visual regions, not general OCR or layout boxes. MP denotes evidence on multiple pages; Contig. additionally requires explicit annotation of continuous support across adjacent page boundaries.

\textbf{Localization and evidence sufficiency.}
Locator summarizes the evidence representation and its annotated hierarchy. XL-DocBench provides pages and supporting text snippets~\citep{xldocbench}; GroundingDocQA-Bench provides nested phrase-, line-, and block-level masks~\citep{m3grounder}; and SciEGQA provides semantically coherent region boxes~\citep{sciegqa}. ReViE adds section context and refines evidence to supporting sentences or necessary figure/table regions, retaining full regions when required. Section titles are actual document headings or \texttt{null}. Refinement seeks precise, collectively sufficient support; multiple valid evidence sets may exist for a question.

\begin{table}[htbp]
    \centering
    \caption{Comparison of ReViE with seven visual document benchmarks. Columns summarize document context, visual evidence, and reasoning and relation supervision.}
    \label{tab:benchmark_comparison}
    \small
    \setlength{\tabcolsep}{3pt}
    \renewcommand{\arraystretch}{1.15}

    \begin{tabular*}{\linewidth}{
        @{\extracolsep{\fill}}lcccccccc@{}
    }
        \toprule

        & \multicolumn{2}{c}{\textbf{Document}}
        & \multicolumn{4}{c}{\textbf{Visual evidence}}
        & \multicolumn{2}{c}{\textbf{Supervision}} \\

        \cmidrule(lr){2-3}
        \cmidrule(lr){4-7}
        \cmidrule(l){8-9}

        \textbf{Benchmark}
        & \textbf{Sci.}
        & \textbf{Input}
        & \textbf{Locator}
        & \textbf{MR}
        & \textbf{MP}
        & \textbf{Contig.}
        & \shortstack{\textbf{Hop}\\\textbf{unit}}
        & \textbf{Rel.} \\

        \midrule

        DocVQA
        & --- & 1 & ---
        & --- & --- & --- & --- & --- \\

        MP-DocVQA
        & --- & Cap & P
        & --- & --- & --- & --- & --- \\

        SlideVQA
        & --- & Cap & P
        & --- & $\checkmark$ & --- & Slide & --- \\

        MMLongBench-Doc
        & $\circ$ & Full & P
        & --- & $\checkmark$ & --- & --- & --- \\

        XL-DocBench
        & $\circ$ & Full & P+T
        & --- & $\checkmark$ & --- & Sect. & --- \\

        GDQA-Bench
        & --- & 1 & M$^{h}$
        & $\checkmark$ & --- & --- & --- & --- \\

        SciEGQA
        & $\checkmark$ & Full & P+B
        & $\checkmark$ & $\checkmark$ & --- & --- & --- \\

        \midrule

        \textbf{ReViE (Ours)}
        & $\checkmark$ & Full & P+S+B
        & $\checkmark$ & $\checkmark$ & $\checkmark$
        & Region & Q/C/I \\

        \bottomrule
    \end{tabular*}

    \smallskip

    \begin{minipage}{\linewidth}
        \footnotesize
        $\checkmark$: explicitly provided;
        $\circ$: scientific content in a mixed-domain corpus;
        ---: not explicitly specified under our criteria.
        MR/MP: multi-region/multi-page evidence;
        Contig.: contiguous cross-page annotation.
        P/S/B/T/M: page/section/box/text snippet/mask;
        $h$: phrase--line--block hierarchy.
        Q/C/I: Quotation/Compression/Inference.
    \end{minipage}
\end{table}

\textbf{Reasoning and relation criteria.}
Hop unit reports the evidence granularity used in explicit multi-hop construction or verification: slides for SlideVQA bridge questions~\citep{slidevqa}, chapter/section branches for XL-DocBench dependency checks~\citep{xldocbench}, and distinct visual regions for ReViE's image-direct multi-hop construction and blind visual verification. Multi-hop status is not inferred from MR, MP, or cross-page continuity alone. Rel. denotes per-evidence support relations, rather than question-level reasoning categories or answer-verification rules. ReViE associates each relation with an evidence item's contribution to the complete answer, without sentence- or claim-level answer decomposition.

\section{Implementation Details}
\label{app:implementation}

\subsection{SFT Configuration}
\label{app:sft_configuration}

VisProve is built on Qwen3-VL-8B~\citep{qwen3vl}. We first perform LoRA-based supervised fine-tuning~\citep{lora} on the 3,115-instance ReViE training split, while keeping the visual backbone frozen. The model is trained to generate a unified serialized target containing the complete answer together with its structured visual provenance.

Let $y=(y_1,\ldots,y_T)$ denote the complete assistant target. The SFT stage optimizes the standard causal language-modeling objective:
\begin{equation}
    \mathcal{L}_{\mathrm{SFT}}
    =
    -\frac{1}{T}
    \sum_{t=1}^{T}
    \log p_{\theta}
    \left(
        y_t \mid D,q,y_{<t}
    \right).
    \label{eq:app_sft}
\end{equation}

We use LoRA with rank 32, scaling factor 64, and dropout 0.05, applied to linear layers of the language model. Optimization uses AdamW~\citep{adamw} with cosine learning-rate scheduling and BF16 precision. The model is trained for three epochs in the main SFT stage with a learning rate of $1\times10^{-4}$. Training then continues for two epochs in the long-document continuation stage with a learning rate of $1\times10^{-5}$. The maximum sequence length is 16,384 and gradient clipping is set to 1.0.

The resulting SFT model serves as the common backbone for all provenance-enhancement variants. Its parameters are frozen during subsequent SER and ERF optimization.

\subsection{SER and ERF Training}
\label{app:ser_erf_training}

SER and ERF are optimized using the same 3,115-instance ReViE training split as SFT, with the SFT backbone kept frozen. All provenance-enhancement configurations are trained for two epochs using AdamW~\citep{adamw} with a learning rate of $1\times10^{-4}$ and weight decay of 0.01. The variants use the same SFT backbone and differ in the optimized components, direct feedback, and associated auxiliary field losses.

\textbf{Spatial visual memory.}
For each document page, frozen visual features are adaptively average-pooled into a $16\times16$ grid containing 256 spatial locations. Each grid feature is normalized, projected to $d=128$ dimensions, and augmented with a two-dimensional positional encoding, yielding the visual-memory vectors $v_{jk}$ used by SER. All document pages participate in the visual memory; no gold evidence page or external retriever is used to preselect the input.

\textbf{Latent bounding-box parameterization.}
Given the SER query $u_t$, the model predicts
\[
    z_t^{b}
    =
    f_{\mathrm{box}}(u_t)
    \in
    \mathbb{R}^{4}.
\]
To ensure an ordered normalized bounding box, we define
\begin{equation}
    l_t
    =
    \sigma(z^{b}_{t,1:2}),
    \qquad
    d_t
    =
    \sigma(z^{b}_{t,3:4}),
    \qquad
    h_t^{b}
    =
    l_t+(1-l_t)\odot d_t,
    \label{eq:app_ser_box}
\end{equation}
and form $\hat b_t=[l_t;h_t^{b}]$ as two ordered corners on the internal $[0,1]^2$ page plane.

SER converts $\hat b_t$ into a differentiable spatial mask over the visual-memory grid. Let $c_{jk}$ denote the center coordinate of location $k$ on page $j$. We compute
\begin{equation}
    \tilde{\alpha}_{tjk}
    =
    \prod_{d\in\{x,y\}}
    \sigma\!\left(
        \kappa(c_{jk,d}-l_{t,d})
    \right)
    \sigma\!\left(
        \kappa(h^{b}_{t,d}-c_{jk,d})
    \right),
    \label{eq:app_soft_mask}
\end{equation}
followed by page-wise normalization:
\begin{equation}
    \alpha_{tjk}
    =
    \frac{
        \tilde{\alpha}_{tjk}
    }{
        \sum_{k'}\tilde{\alpha}_{tjk'}+\epsilon
    }.
    \label{eq:app_soft_mask_norm}
\end{equation}
We set $\kappa=30$, which controls the sharpness of the differentiable region boundary.

\textbf{Norm-bounded residual feedback.}
For the representation $z_t$ defined in Equation~(\ref{eq:module_combination}), we first project it to the decoder dimension:
\begin{equation}
    \delta_t
    =
    W_{\mathrm{up}}z_t.
    \label{eq:app_residual_projection}
\end{equation}
Its magnitude is bounded relative to the original hidden state:
\begin{equation}
    \lambda_t
    =
    \min
    \left(
        1,\,
        \frac{
            0.03\lVert h_t\rVert_2
        }{
            \max(\lVert\delta_t\rVert_2,\epsilon)
        }
    \right).
    \label{eq:app_residual_scale}
\end{equation}
A learned gate further controls the residual contribution:
\begin{equation}
    g_t
    =
    \sigma(W_gu_t+b_g),
    \qquad
    \Delta h_t
    =
    g_t\lambda_t\delta_t,
    \qquad
    h'_t
    =
    h_t+\Delta h_t.
    \label{eq:app_gated_residual}
\end{equation}
Thus, before gating, the residual norm is bounded by $3\%$ of $\lVert h_t\rVert_2$, while $g_t$ may further reduce its contribution. The modification is applied after the final decoder layer and before the original output head.

\textbf{Training objective.}
The objective in Equation~(\ref{eq:overall_objective}) uses the following weights:
\begin{equation}
\begin{aligned}
    \mathcal{L}
    ={}&
    \mathcal{L}_{\mathrm{gen}}
    +
    0.1
    \left[
        \mathbb{I}_{\mathrm{SER}}
        \left(
            \mathcal{L}_{\mathrm{page}}
            +
            \mathcal{L}_{\mathrm{box}}
        \right)
        +
        \mathbb{I}_{\mathrm{ERF}}
        \mathcal{L}_{\mathrm{rel}}
    \right] \\
    &+
    0.25\mathcal{L}_{\mathrm{field}}
    +
    10\mathcal{L}_{\mathrm{outside}}
    +
    0.1\mathbb{I}_{\mathrm{SER}}
    \mathcal{L}_{\mathrm{gate}}.
\end{aligned}
\label{eq:app_full_objective}
\end{equation}

\textbf{Supervision alignment and trainable components.}
Auxiliary targets are paired with gold evidence items in their serialized order after checking that the field counts agree. Field-value character spans are mapped to tokenizer positions. Let $\tau_b$, $\tau_s$, and $\tau_r$ denote the decoder positions immediately preceding the first coordinate token of \texttt{rel\_bbox}, the first content token of \texttt{section}, and the first content token of \texttt{relation}, respectively. Table~\ref{tab:auxiliary_supervision} summarizes the supervision positions and trainable components. In \emph{w/o SER}, the frozen spatial reader still supplies $s_t$ to ERF, but only $e_t$ is directly injected. Serialized page values are learned through generation cross-entropy, and there is no separate section-title classification head.

\begin{table}[htbp]
\centering
\caption{Configuration-specific training of provenance-enhancement components. The SFT backbone remains frozen in all variants.}
\label{tab:auxiliary_supervision}
\footnotesize
\setlength{\tabcolsep}{4pt}
\renewcommand{\arraystretch}{1.12}
\begin{tabular}{
    @{}
    >{\raggedright\arraybackslash}p{0.15\linewidth}
    >{\raggedright\arraybackslash}p{0.28\linewidth}
    >{\raggedright\arraybackslash}p{0.28\linewidth}
    >{\raggedright\arraybackslash}p{0.21\linewidth}
    @{}
}
\toprule
\textbf{Configuration}
& \textbf{Trainable components}
& \textbf{Auxiliary supervision}
& \textbf{Field CE values} \\
\midrule

\emph{w/o ERF}
& Spatial reader, geometric fusion, gate, and up-projection
& Page and box at $\tau_b$, with two loss contributions
& \texttt{rel\_bbox} \\

\addlinespace[0.4em]

\emph{w/o SER}
& Relation classifier and relation embeddings only
& Relation at $\tau_r$
& \texttt{section}, \texttt{relation} \\

\addlinespace[0.4em]

VisProve
& All provenance-enhancement components
& Page, box, and relation at $\tau_r$; page and box at $\tau_s$
& \texttt{rel\_bbox}, \texttt{section}, \texttt{relation} \\

\bottomrule
\end{tabular}
\end{table}

\textbf{Loss computation and reductions.}
$\mathcal{L}_{\mathrm{gen}}$ is causal cross-entropy averaged over the complete answer--provenance target and its end-of-sequence token. Each page or relation classification loss is averaged over evidence items; each box loss combines mean coordinate-level $\ell_1$ error with mean $1-\mathrm{IoU}$ over evidence items. $\mathcal{L}_{\mathrm{page}}$ and $\mathcal{L}_{\mathrm{box}}$ sum the corresponding contributions listed in Table~\ref{tab:auxiliary_supervision}, without averaging over the number of contributions. In \emph{w/o ERF}, the section-associated localization call is mapped to $\tau_b$, so both page-and-box contributions are retained even though they use the same position.

\textbf{Field masks and residual regularization.}
$\mathcal{L}_{\mathrm{field}}$ averages causal cross-entropy over tokens overlapping the field-value spans listed in Table~\ref{tab:auxiliary_supervision}, including JSON \texttt{null} for unresolved sections and punctuation within the selected spans. Both generation and field losses supervise the prediction of each token from its preceding decoder state. The field-token mask also defines positive gate targets, with other output positions assigned negative targets. $\mathcal{L}_{\mathrm{gate}}$ uses binary cross-entropy with equal weight for the positive and negative groups and is included only for \emph{w/o ERF} and VisProve. $\mathcal{L}_{\mathrm{outside}}$ averages the relative residual energy
\begin{equation}
    \frac{
        \lVert\Delta h_t\rVert_2^2
    }{
        \lVert h_t\rVert_2^2
    }
    \label{eq:app_outside}
\end{equation}
over decoder positions predicting output tokens outside the selected fields. These positions include answer text and field names, which remain supervised by $\mathcal{L}_{\mathrm{gen}}$. Prompt tokens do not contribute to the output losses.

\subsection{Inference Configuration}
\label{app:inference_configuration}

At inference time, the model receives the question together with all pages of the corresponding document in their original order. Gold evidence pages, sections, bounding boxes, and relation labels are not provided. We use greedy decoding with a maximum of 768 newly generated tokens, without sampling or beam search. The answer and all provenance fields are generated autoregressively by the original language head.

For SER and ERF, the per-page visual budget is adjusted according to document length to control the total visual context: documents with at most 12 pages use up to 262,144 pixels per page, documents with 13--32 pages use up to 131,072 pixels per page, and longer documents use up to 65,536 pixels per page.

No post-hoc bounding-box calibration, section lookup, relation rewriting, or other prediction repair is applied. The latent bounding box predicted by SER is used only for internal differentiable visual reading and does not overwrite the bounding box generated in the final structured output.

The decoding and visual-budget settings in this subsection apply to VisProve and its two feedback variants. General-purpose baselines use the shared task prompt in Appendix~\ref{app:inference_prompt} with model-specific image processing, while VISA, CoE, and UniDoc-RL follow their native inference procedures as described in Section~\ref{sec:method_comparison}.

\subsection{Unified Inference Prompt}
\label{app:inference_prompt}

Figure~\ref{fig:inference_prompt} shows the unified task prompt used for general-purpose open- and closed-source VLM baselines and for the SFT/SER/ERF variants. Each instance additionally provides all document pages in their original order together with the question. Reference answers and gold provenance annotations are not included at inference time. Specialized methods in Section~\ref{sec:method_comparison} follow their respective native inference procedures.

\begin{promptfigure}{Unified Inference Prompt}
\small

You are a scientific document question-answering assistant with visual provenance grounding.

The input images together form a single scientific document and are provided in page order. Answer the question using only the information contained in the document.

You must output both the answer and the minimum sufficient visual evidence that supports the answer. The evidence array must not be empty.

\textbf{For each evidence item:}

\begin{itemize}
    \setlength{\itemsep}{0pt}
    \setlength{\parskip}{0pt}

    \item \texttt{"page"} is the 1-based position of the corresponding page in the input image sequence.

    \item \texttt{"section"} must be the actual section or subsection title shown in the document. If it cannot be reliably determined, use the JSON value \texttt{null}, not the string \texttt{"null"}.

    \item \texttt{"rel\_bbox"} must contain four numeric values in the range 0--1000, formatted as \texttt{[x1, y1, x2, y2]}, where $x1<x2$ and $y1<y2$. Coordinates may be integers or decimals; do not assume they must be integers.

    \item For textual evidence, the bounding box must cover the complete supporting sentence or sentences.

    \item For figures and tables, use the minimum sufficient visual region that supports the answer. Use the full figure or table only when necessary.

    \item \texttt{"relation"} must be exactly one of \texttt{"Quotation"}, \texttt{"Compression"}, or \texttt{"Inference"}.

    \item \texttt{"relation"} describes how this specific evidence item supports the complete answer. When multiple evidence items are provided, label each item separately.
\end{itemize}

\textbf{Relation definitions:}

\begin{itemize}
    \setlength{\itemsep}{0pt}
    \setlength{\parskip}{0pt}

    \item \textbf{Quotation:} the relevant answer information can be directly read from the evidence or is only lightly rephrased.

    \item \textbf{Compression:} the answer faithfully summarizes, condenses, or combines information explicitly present in the evidence without introducing a new logical conclusion.

    \item \textbf{Inference:} the answer requires comparison, calculation, ordering, logical deduction, conditional reasoning, or integration of evidence to derive a conclusion that is not directly stated.
\end{itemize}

Return strictly valid JSON with exactly two top-level fields, \texttt{"answer"} and \texttt{"evidence"}. Each evidence item must contain exactly \texttt{"page"}, \texttt{"section"}, \texttt{"rel\_bbox"}, and \texttt{"relation"}.

Do not output Markdown, explanations, comments, or any text outside the JSON object.

\end{promptfigure}

\captionof{figure}{Unified inference prompt used for general-purpose VLM baselines and VisProve variants. The prompt specifies complete-document input, structured answer--provenance output, visual localization requirements, and evidence--answer relation types.}
\label{fig:inference_prompt}

\section{Evaluation Details}
\label{app:evaluation}

\subsection{Metric Definitions}
\label{app:metric_definitions}

We evaluate models along three complementary dimensions. \emph{Answer quality} is measured by ROUGE-L and BERTScore-F1. \emph{Evidence localization} is evaluated using Page-F1, Section-F1, Evidence-IoU, and Region-F1, covering page identification, section localization, and fine-grained visual-region overlap. \emph{Typed provenance} is measured by Relation-F1 and Typed-F1, which assess relation-consistent evidence prediction and complete typed provenance, respectively. All task-specific metrics are reported on a 0--100 scale, with higher values indicating better performance.

\textbf{Answer quality.}
ROUGE-L is computed between the predicted and reference answers after basic Unicode and whitespace normalization. BERTScore-F1 measures their semantic similarity using contextual representations. Both metrics are evaluated over all 409 test instances.

\textbf{Page and section localization.}
Page-F1 treats the predicted and reference page indices as sets and computes
\begin{equation}
    \mathrm{Page\!-\!F1}
    =
    \frac{
        2|\hat{\Pi}\cap\Pi|
    }{
        |\hat{\Pi}|+|\Pi|
    },
\end{equation}
where $\hat{\Pi}$ and $\Pi$ denote the predicted and reference evidence-page sets, respectively. Page-F1 is averaged over all test instances.

Section-F1 similarly compares the sets of predicted and reference non-\texttt{null} section names. Section strings are normalized using Unicode normalization, case folding, whitespace normalization, and removal of leading numerical section identifiers. Section-F1 is evaluated only on instances containing at least one non-\texttt{null} reference section; a missing predicted section receives zero credit for such an instance.

\textbf{Region and typed provenance metrics.}
Evidence-IoU, Region-F1, Relation-F1, and Typed-F1 use one-to-one matching between predicted and reference evidence regions. Bounding boxes are evaluated directly in the normalized coordinate space, and region similarity is measured by continuous IoU rather than thresholding at a fixed IoU value. The detailed matching procedure is described in Appendix~\ref{app:evidence_matching}.

We additionally report a text-only LLM-as-Judge score on a shared 50-instance subset of the test set. The judge assigns an integer score from 0 to 5 to the predicted answer. This score is reported separately and is not included in the Overall metric.

\subsection{Evidence Matching and Failure Handling}
\label{app:evidence_matching}

Let $P$ and $G$ denote the numbers of predicted and reference evidence items for an instance. For region-based metrics, we construct pairwise IoU scores between predicted and reference regions and perform maximum-weight one-to-one matching using the Hungarian algorithm. Unmatched predicted or reference evidence remains in the corresponding denominator, penalizing both over-prediction and missing evidence.

\textbf{Evidence-IoU.}
Predicted and reference regions are matchable only when they occur on the same page. For instances whose predicted page set exactly matches the reference page set, Evidence-IoU is computed as
\begin{equation}
    \mathrm{Evidence\!-\!IoU}
    =
    \frac{
        \sum_{(i,j)\in\mathcal{M}} \mathrm{IoU}(b_i,\hat b_j)
    }{
        \max(P,G)
    },
\end{equation}
where $\mathcal{M}$ is the maximum-weight one-to-one matching. The final Evidence-IoU is averaged over test instances satisfying the page-set exact-match condition.

\textbf{Region-F1.}
Using the same page-consistent IoU matching, Region-F1 is
\begin{equation}
    \mathrm{Region\!-\!F1}
    =
    \frac{
        2\sum_{(i,j)\in\mathcal{M}}\mathrm{IoU}(b_i,\hat b_j)
    }{
        P+G
    }.
\end{equation}
Region-F1 is likewise averaged over instances whose predicted and reference page sets are identical.

\textbf{Relation-F1.}
Relation-F1 additionally requires the predicted and reference evidence items to have the same relation label. Only same-page, relation-consistent pairs retain their IoU score before Hungarian matching:
\begin{equation}
    \mathrm{Relation\!-\!F1}
    =
    \frac{
        2\sum_{(i,j)\in\mathcal{M}_{\mathrm{rel}}}
        \mathrm{IoU}(b_i,\hat b_j)
    }{
        P+G
    }.
\end{equation}
Like Evidence-IoU and Region-F1, Relation-F1 is averaged over page-exact test instances. This metric evaluates relation-consistent evidence matching rather than relation classification in isolation.

\textbf{Typed-F1.}
Typed-F1 applies the strongest matching condition. A matched pair must agree on page, section, and evidence--answer relation, in addition to its continuous visual overlap:
\begin{equation}
    \mathrm{Typed\!-\!F1}
    =
    \frac{
        2\sum_{(i,j)\in\mathcal{M}_{\mathrm{typed}}}
        \mathrm{IoU}(b_i,\hat b_j)
    }{
        P+G
    }.
\end{equation}
When the reference section is \texttt{null}, the section condition is treated as satisfied; when the reference section is non-\texttt{null}, the normalized predicted and reference section names must match. Unlike Evidence-IoU, Region-F1, and Relation-F1, Typed-F1 is evaluated over all 409 test instances rather than conditioning on exact page-set prediction.

\textbf{Invalid and incomplete predictions.}
Predicted page indices outside the valid document range are discarded. A retained evidence item with an invalid or missing bounding box receives zero region overlap while remaining in the prediction count, and an invalid or missing relation label cannot satisfy relation-aware matching. Partially malformed structured outputs are evaluated using recoverable fields only; gold annotations are never used to repair model predictions.

\subsection{Overall Score}
\label{app:overall_score}

\textbf{Main evaluation.}
For the main results, we first average metrics within the three evaluation dimensions:
\begin{equation}
\begin{aligned}
    A &= \frac{\mathrm{ROUGE\!-\!L}+\mathrm{BERTScore\!-\!F1}}{2},\\
    L &= \frac{\mathrm{Page\!-\!F1}+\mathrm{Section\!-\!F1}
    +\mathrm{Evidence\!-\!IoU}+\mathrm{Region\!-\!F1}}{4},
\end{aligned}
\end{equation}
and
\begin{equation}
    T
    =
    \frac{
        \mathrm{Relation\!-\!F1}
        +
        \mathrm{Typed\!-\!F1}
    }{2},
    \qquad
    \mathrm{Overall}
    =
    \frac{A+L+T}{3}.
    \label{eq:overall_metric}
\end{equation}
This formulation gives equal weight to answer quality, evidence localization, and typed provenance regardless of the number of individual metrics in each category.

\textbf{Existing-method comparison.}
VISA, CoE, and UniDoc-RL do not provide the section and typed-relation outputs required by the full ReViE evaluation. We therefore evaluate their shared answer and localization metrics using the same definitions as in the main evaluation, but compute a comparison-specific Overall score over only these shared dimensions:
\begin{equation}
    A_{5}
    =
    \frac{
        \mathrm{ROUGE\!-\!L}
        +
        \mathrm{BERTScore\!-\!F1}
    }{2},
    \qquad
    L_{5}
    =
    \frac{
        \mathrm{Page\!-\!F1}
        +
        \mathrm{Evidence\!-\!IoU}
        +
        \mathrm{Region\!-\!F1}
    }{3},
\end{equation}
\begin{equation}
    \mathrm{Overall}_{\mathrm{shared}}
    =
    \frac{A_{5}+L_{5}}{2}.
    \label{eq:method_overall}
\end{equation}
This comparison-specific score is used only in Table~\ref{tab:method_comparison} and is not directly comparable to the three-category Overall score used in Table~\ref{tab:main_results}.

\subsection{LLM-as-Judge Protocol}
\label{app:judge_protocol}

We use GPT-5.6 Luna as a blind, text-only judge on a shared 50-instance subset of the 409-instance ReViE test set. All evaluated systems are scored on the same set of questions. For each instance, the judge receives only the \emph{Question}, \emph{Reference Answer}, and \emph{Predicted Answer}; document pages, evidence regions, section annotations, relation labels, provenance predictions, and model identities are not provided.

The judge evaluates the predicted answer jointly in terms of correctness, completeness, and faithfulness to the reference answer. It assigns a single integer score from 0 to 5 according to the following rubric:
\begin{itemize}
    \item \textbf{5}: fully correct, complete, and faithful, with no material error or omission;
    \item \textbf{4}: the core answer is correct, with only a minor omission, imprecision, or wording issue;
    \item \textbf{3}: substantial correct content is present, but with one major error or omission, or several minor issues;
    \item \textbf{2}: some relevant content is correct, but key information is missing or materially incorrect;
    \item \textbf{1}: only weakly relevant content is present, with little usable answer to the question;
    \item \textbf{0}: the response is completely incorrect, contradictory, off-topic, empty, a refusal, or otherwise unusable.
\end{itemize}

Faithful paraphrases, equivalent scientific notation or units, and harmless rounding are accepted, and lexical overlap is not required. Each system's Judge score is the arithmetic mean over the shared 50 evaluated instances. The judge evaluates answer content only and does not inspect visual evidence or provenance. Its score is therefore treated as an auxiliary answer-quality signal rather than a provenance metric and is excluded from Overall.

\textbf{Judge prompt.}
The complete system prompt and user template used for LLM-based answer
evaluation are shown in Figure~\ref{fig:judge_prompt}.

\begin{promptfigure}{LLM-as-Judge Prompt}
\small

You are the fixed, blind, text-only answer evaluator for ReViE, a scientific-document question-answering benchmark.

\textbf{INPUT BOUNDARY}

You receive exactly three fields: Question, Reference Answer, and Predicted Answer. The Reference Answer is the authoritative evaluation target. Use no external facts. Never evaluate or infer pages, images, bbox coordinates, sections, evidence, provenance, model identity, or writing style beyond whether the answer is clear enough to assess.

\textbf{SCORING PRINCIPLES}

Evaluate the Predicted Answer jointly for:

\begin{enumerate}
    \setlength{\itemsep}{0pt}
    \setlength{\parskip}{0pt}
    \item \textbf{Correctness:} its substantive claims agree with the Reference Answer.
    \item \textbf{Completeness:} it covers every part needed to answer the Question.
    \item \textbf{Faithfulness:} it does not add material claims contradicted by, or unsupported by, the Reference Answer.
\end{enumerate}

Accept faithful paraphrases, synonyms, reordered statements, equivalent scientific notation, equivalent units, and harmless rounding that preserves the meaning. Do not score by lexical overlap. Do not require an explanation when a short answer fully answers the Question. For multi-part questions, assess every requested part. Treat a material reversal, wrong entity/value, or contradiction as a major error. An empty answer, refusal, or unusable output scores 0.

\textbf{INTEGER RUBRIC}

\begin{itemize}
    \setlength{\itemsep}{0pt}
    \setlength{\parskip}{0pt}
    \item \textbf{5}: Fully correct, complete, and faithful; no material error or omission.
    \item \textbf{4}: Core answer is correct; only a minor omission, imprecision, or wording issue.
    \item \textbf{3}: Substantial correct content, but one major omission/error or several minor ones.
    \item \textbf{2}: Some relevant correct content, but key information is missing or materially wrong.
    \item \textbf{1}: Only weakly relevant content; almost no usable answer to the Question.
    \item \textbf{0}: Completely incorrect, contradictory, off-topic, empty, refusal, or unusable.
\end{itemize}

\textbf{OUTPUT CONTRACT}

Return JSON only, with exactly:

{\ttfamily\footnotesize
\{"score": <integer 0..5>, "reasoning": "<brief auditable justification>"\}
}

The justification must identify the decisive agreement, omission, or error in one or two concise sentences. Do not output markdown or additional keys.

\textbf{USER MESSAGE TEMPLATE}

Evaluate this answer.

\textbf{Question:}\\
\texttt{\{question\}}

\textbf{Reference Answer:}\\
\texttt{\{reference\_answer\}}

\textbf{Predicted Answer:}\\
\texttt{\{predicted\_answer\}}

\end{promptfigure}

\captionof{figure}{Prompt used for blind text-only LLM-as-Judge evaluation. The judge receives only the question, reference answer, and predicted answer and assigns an integer score from 0 to 5.}
\label{fig:judge_prompt}

\subsection{Human Evaluation Protocol}
\label{app:human_evaluation}

We conduct human evaluation on a shared subset of 20 test questions across 11 models, yielding 220 model--question answer observations. The evaluated systems are Qwen3-VL-8B Base, InternVL3.5-8B, MiniCPM-V4.5-8B, Qwen3-VL-32B, InternVL3.5-38B, Qwen2.5-VL-7B, Eagle2.5-8B, GPT-4o-mini, Claude Haiku 4.5, Gemini 2.5 Flash-Lite, and VisProve. Ablation variants are not included in this analysis.

Four expert reviewers participate in the evaluation. Reviewers assess answers independently using only the \emph{Question}, \emph{Reference Answer}, and \emph{Predicted Answer}. Document pages, separately provided evidence or provenance annotations, model identities, and automated evaluation scores are not shown. Although provenance-like content may appear in a model's raw output, no separate evidence fields are explicitly provided to the reviewers.

Each reviewer assigns an integer score from 0 to 5 based jointly on correctness, completeness, and faithfulness to the reference answer. Here, faithfulness is assessed against the reference answer; reviewers do not inspect the source document. The scoring rubric is:
\begin{itemize}
    \item \textbf{5}: fully correct and complete, with no material error or omission;
    \item \textbf{4}: the core answer is correct, with only minor issues;
    \item \textbf{3}: substantial correct content is present, but with a major omission or error;
    \item \textbf{2}: some relevant content is correct, but key information is missing or materially incorrect;
    \item \textbf{1}: the response is only weakly relevant and contains little usable answer content;
    \item \textbf{0}: the response is incorrect, off-topic, empty, or otherwise unusable.
\end{itemize}

All 220 answers are independently scored by three reviewers. A fourth reviewer independently re-evaluates 16 answers exhibiting larger disagreement among the initial three reviewers. No discussion-based consensus score is used. For the 16 re-evaluated answers, the final human score is the median of four ratings; for all remaining answers, it is the median of three ratings. Consequently, the final per-answer human score may take half-point values in the re-evaluated cases.

For consistency analysis, we compute each model's human score as the arithmetic mean over the same 20 questions. The LLM-as-Judge score is computed over these same model--question pairs, and each point in Figure~\ref{fig:human_consistency} represents one model's mean human score and mean LLM-as-Judge score. Pearson's correlation across the 11 model-level pairs is $r=0.9898$. The correlation is computed over 11 model-level pairs; instance-level agreement is a separate quantity.
\begin{figure}[t]
    \centering
    \includegraphics[width=\linewidth]{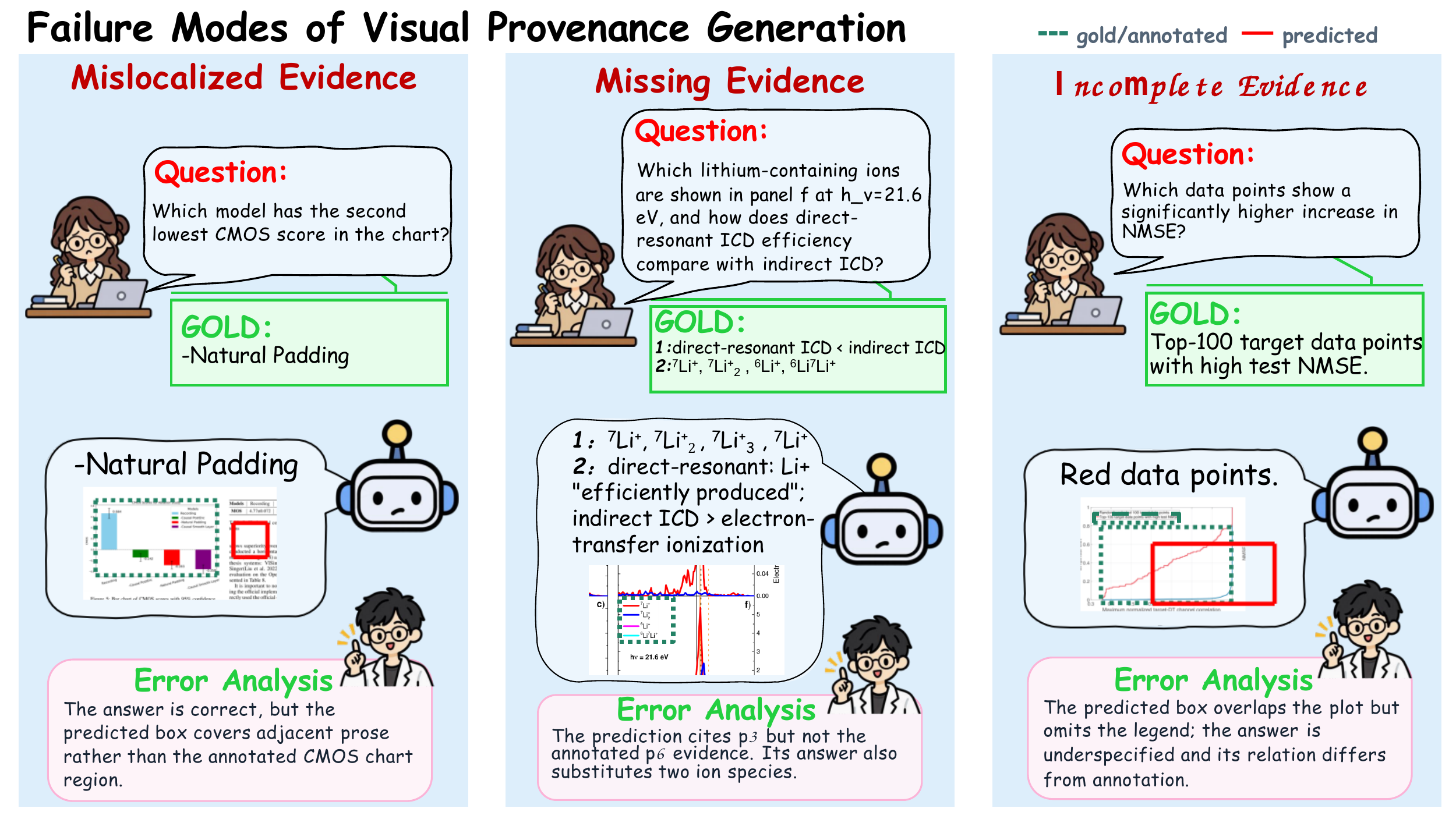}
    \caption{Representative prediction errors. Left: a correct chart-reading answer paired with a misplaced evidence box. Middle: incomplete cross-page evidence coverage accompanied by an incorrect identification. Right: incomplete evidence coverage, an underspecified answer, and a relation discrepancy.}
    \label{fig:failure_cases}
\end{figure}

\section{Qualitative Error Analysis}
\label{app:qualitative_error_analysis}

Figure~\ref{fig:failure_cases} presents three representative failure patterns observed in VisProve predictions.

\textbf{Region localization mismatch.}
The left case answers the CMOS ranking question correctly, but places the evidence box over adjacent prose. The reported region differs from the chart region supporting the answer.

\textbf{Cross-page evidence omission.}
The middle case requires introductory text and a later figure. The prediction cites only the introductory page and misidentifies two ion species. Its output exhibits both incomplete evidence coverage and an answer error.

\textbf{Evidence coverage and relation discrepancy.}
In the right case, a predicted box overlaps an annotated plot region but omits a separate legend region. The answer gives a color-based description with insufficient specificity, and its predicted relation differs from the annotation. This case combines evidence-coverage, answer-completeness, and relation discrepancies.

These examples distinguish several forms of disagreement with annotated support. Their interpretation is relative to the reference evidence sets, which may coexist with other valid support.

\end{document}